\documentclass{article}

\usepackage[numbers]{natbib}

\usepackage {arxiv}

\usepackage{cite}
\usepackage{floatrow}
\usepackage{amsmath,amssymb,amsfonts}
\usepackage{algorithm}
\usepackage{algorithmic}
\usepackage{textcomp}
\usepackage{xcolor}
\usepackage{float}
\usepackage{booktabs}
\usepackage{multirow}
\usepackage{enumitem}
\usepackage{setspace}
\usepackage[]{graphicx}
\usepackage{array}
\usepackage{amsmath}
\usepackage{amssymb}
\usepackage{epsfig}
\usepackage{epstopdf}
\usepackage{subfigure}
\usepackage{comment}
\usepackage{tabularx}
\usepackage{threeparttable}
\usepackage{adjustbox}
\usepackage{soul}
\usepackage{dsfont}
\usepackage{hyperref} 

\def\BibTeX{{\rm B\kern-.05em{\sc i\kern-.025em b}\kern-.08em
    T\kern-.1667em\lower.7ex\hbox{E}\kern-.125emX}}

\title{\LARGE \bf
DeepIFSAC: Deep Imputation of Missing Values Using Feature and Sample Attention within Contrastive Framework}

\author{Ibna Kowsar,~
        Shourav B. Rabbani, ~
        Yina Hou,~
        Manar D. Samad\\
Department of Computer Science\\
Tennessee State University\\
Nashville, TN, USA\\
\texttt{msamad@tnstate.edu} \\
}
 
\begin{document}

\maketitle

\begin{abstract}

Missing values of varying patterns and rates in real-world tabular data pose a significant challenge in developing reliable data-driven models. The most commonly used statistical and machine learning methods for missing value imputation may be ineffective when the missing rate is high and not random. This paper explores row and column attention in tabular data as between-feature and between-sample attention in a novel framework to reconstruct missing values. The proposed method uses CutMix data augmentation within a contrastive learning framework to improve the uncertainty of missing value estimation. The performance and generalizability of trained imputation models are evaluated in set-aside test data folds with missing values. The proposed framework is compared with 11 state-of-the-art statistical, machine learning, and deep imputation methods using 12 diverse tabular data sets. The average performance rank of our proposed method demonstrates its superiority over the state-of-the-art methods for missing rates between 10\% and 90\% and three missing value types, especially when the missing values are not random. The quality of the imputed data using our proposed method is compared in a downstream patient classification task using real-world electronic health records. This paper highlights the heterogeneity of tabular data sets to recommend imputation methods based on missing value types and data characteristics.

\end{abstract}


\section{Introduction}

In the era of big data-driven intelligence, meaningful insights from large data sets are crucial in advancing science, the economy, and industries.  The quality and utility of data are often challenged due to the presence of missing values at varying rates and patterns in many domains, including electronic health records (EHR), banking, biology, and cybersecurity~\citep{Kazijevs2023, Grinsztajn2022, Borisov2022, Kadra2021}. Missing values in the row-column structure of tabular data are mainly studied in statistics. Moreover, traditional machine learning (ML) remains the preferred approach for tabular data, despite the revolutionary feat of deep learning (DL) in artificial intelligence (AI) for image and text data~\citep{Kadra2021, Grinsztajn2022, FTT_Gorishniy2021}. DL/ML methods are often benchmarked using clean and curated data, whereas real-world data are rarely ready for DL/ML due to missing values. For example, EHR data inevitably contain missing values because physicians never recommend all tests for every patient or the same test in all medical follow-ups. However, imputation of missing values to enable DL / ML can alter data quality and impact data-driven results. Therefore, missing value imputation (MVI) methods aim to prepare high-quality and unbiased data sets for downstream predictive modeling. 

The recent literature has presented statistical, machine learning, and deep learning methods for MVI. A common approach is to estimate missing values using feature regression models~\citep{mice, missForest}, feature correlation~\citep{khan2024high, khan2022missing}, feature attention~\citep{aimnet, lee2023}, and generative models~\citep{yoon2018gain, ouyang2023missdiff}. We argue that a missing value at the intersection of a row and a column should be estimated by jointly learning column-attention and row-attention to fully leverage the row-column structure of tabular data. Existing column attention methods, namely self-attention, learn from feature dependencies, ignoring row attention to learn missing values from similar samples. Our proposed Deep Imputation using Feature and Sample Attention within the Contrastive (DeepIFSAC) framework adds complementary values to self-attention from between-sample dependencies using row attention. When feature dependencies are weak, missing value estimation may benefit from attending to similar samples via row attention. The proposed missing value estimation is further enhanced by jointly learning column attention and row attention in a contrastive learning framework. Contrastive learning yields a deep representation by keeping similar samples together and dissimilar samples farther away, which aids row attention in minimizing the contribution of outliers or dissimilar samples in MVI. The source code for the implementation of the proposed algorithm has been publicly shared\footnote{github link would be shared in the final version for anonymity during the peer-review process}. To the best of our knowledge, this paper is the first to propose a joint row-column attention augmented by contrastive learning for missing value estimation in tabular data. The contributions of this paper are as follows.
\begin{itemize}
\item The proposed MVI method combines between-feature and between-sample attention to learn between-column and between-row dependencies of the row-column structure of tabular data.
\item The robustness of MVI is enhanced by the CutMix data augmentation method~\citep{cutmix_yun2019} during model training instead of filling missing values with zero or median values~\citep{Midaspy_Lall_2022, lee2023}. 
\item Contrastive learning attracts similar samples closer and dissimilar ones farther apart in deep representation and facilitates MVI attending to similar samples, minimizing the contributions of outliers.
\item Unlike other studies, we perform a thorough benchmarking using 12 tabular data sets with varying characteristics, three missing value types, and missing rates between 10\% and 90\%. The quality of the imputed data is evaluated using real electronic health records to classify patients with heart failure. 
\end {itemize}

The structure of the paper is organized in the following order. Section 2 reviews the state-of-the-art traditional and deep learning methods for MVI. The steps of the proposed joint attention learning framework are presented in Section 3. Section 4 discusses the data sets, the experimental setup and scenarios, and the evaluation method. Section 5 presents the imputation performance of our method and compares it with other leading baseline MVI methods. Section 6 summarizes the main findings, provides insight into the results, and outlines limitations with possible avenues for future investigation. The paper concludes in Section 7.

\section{Literature review}
\label{background}

Multiple imputations (MI) are one of the pioneering statistical approaches to MVI, which is the foundation of multiple imputations using chained equations (MICE)~\citep{mice}, missForest~\citep{missForest}, and deep MICE~\citep{Samad2022}. MICE iteratively updates missing value estimates for individual variables using linear regression models. Nonlinear machine learning regressions, such as random forests and gradient-boosting trees, have replaced linear regression and demonstrated notable improvements in imputation accuracy~\citep{missForest, dimitrispredictive, Samad2022}. However, regression models can produce values out of range. Nonlinear models are known to overfit data with limited samples and high dimensionality of features~\citep{emmanuel2021survey}. Mera et al. ~\citep{Mera2021} argue that MICE is computationally expensive in complex data scenarios. The number and complexity of the regression equations increase with the feature dimension. 

On the other hand, deep learning methods have rapidly evolved and revolutionized artificial intelligence (AI), primarily for image and text data. However, deep learning solutions~\citep{Kadra2021, Borisov2022, FTT_Gorishniy2021, Chen2023} proposed for tabular data have not yet replaced their competitive counterparts in machine learning. Kadra et al.~\citep{Kadra2021} refer to tabular data as the ``last unconquered castle" for deep learning. The limited success of deep learning on tabular data has inspired recent research on deep imputation methods. In ~\citep{Samad2022}, deep regression within the MICE framework has shown promising performance in imputing up to 80\% missing values. The study shows that deep regression within MICE is superior to its machine learning counterparts for imputing missing not at random (MNAR) type data. In contrast, generative deep learning has inspired several missing value imputation models. The diffusion-based framework (MissDiff) is trained to generate values for missing data entries~\citep{ouyang2023missdiff}. Diffusion models may be suboptimal for tabular data imputation due to their reliance on a Gaussian noise process, which restricts their ability to accurately capture complex patterns in missing values~\citep{rethinking_diff}. A generative model, Generative Adversarial Imputation Nets (GAIN)~\citep{yoon2018gain}, has been evaluated using 20\% completely at random missing values. However, GAIN assumes that missing values are completely at random (MCAR), which may limit its application to missing value patterns that are not completely at random~\citep{diffimpute}. For example, Sun et al. argue that GAIN has limited performance with MNAR and missing at random (MAR) data types, where MICE and MissForest often outperform GAIN~\citep{sun2023deep}. Furthermore, generative adversarial networks (GANs) can be unstable during optimization, particularly in terms of model convergence\citep{hyperimpute}. Missing values are also reconstructed from the deep representation of self-supervised autoencoder models. Using the multiple-imputation (MI)-based denoising autoencoder (DA) method, MIDASpy~\citep{Midaspy_Lall_2022} replaces a random subset of training data with zeros during each mini-batch processing. However, initializing missing values with zeros can substantially alter the original data distribution and eventually affect the performance of the model~\citep{liguori2023augmenting}.  The performance of MIDASpy is also unknown compared to similar autoencoder-based imputation methods~\citep{MIDA_Gondara2018,AE2}.

Recent MVI methods take inspiration from state-of-the-art deep learning models. The convolution operation of deep convolutional neural networks (CNNs) is leveraged in MVI by transforming tabular data into a pseudo-spatial 2D grid-like structure~\citep{khan2024high, khan2022missing}. Notably, CNN is highly effective for image data due to spatial regularity in feature space, unlike the permutation-invariant feature space of tabular data without such regularity. Therefore, deep learning methods tailored to tabular data have recently emerged. Multiple attention and transformer-based models have been proposed for tabular data classification, including the Feature Tokenizer Transformer (FT-Transformer)~\citep{FTT_Gorishniy2021} and the Non-parametric Transformer (NPT)~\citep{NPT_Kossen2021}. A recent benchmarking article \citep{rabbani2024} suggests that combining deep contrastive and attention learning, for example, Self-Attention and Intersample Attention Transformer (SAINT)~\citep{SAINT_Somepalli2022}, significantly outperforms most baselines in classifying a wide selection of tabular data. However, it is unknown whether similar deep representation learning methods can be repurposed for missing value imputation problems. 

Attention between features (aka self-attention) can be effective when missing feature values depend on other features. A denoising self-attention network (DSAN)~\citep{lee2023} has been proposed for MVI. Like~\citep{Midaspy_Lall_2022}, a random subset of the training data is replaced with zeros during the mini-batch training of the network. DSAN shows limited performance on up to 20\% of completely random missing values. Similarly, the attention-based imputation network (AimNet) learns missing values from attention between features only~\citep{aimnet}. Recent studies have proposed data clustering to improve MVI performance by taking contributions from similar samples within a cluster~\citep{Samad2022, khan2022handling}. In addition to learning feature dependencies, similarity or dependency between samples may provide complementary values to MVI.   

Considering the gaps in the literature, this paper proposes a novel deep MVI method that jointly learns between-feature and between-sample attention for the column-row structure of tabular data. A contrastive learning framework with data augmentation improves deep representation, facilitating attention between samples for robust MVI. A rigorous analysis of imputation methods on three missing value types from 10\% to 90\% missing rates using a diverse set of tabular data demonstrates the effectiveness of proposed imputation methods against state-of-the-art MVI methods.

\section {Methods}

\begin{figure*}[t]
\centering
\includegraphics[trim=0cm 0cm 0cm 0cm, width=1.0\textwidth] {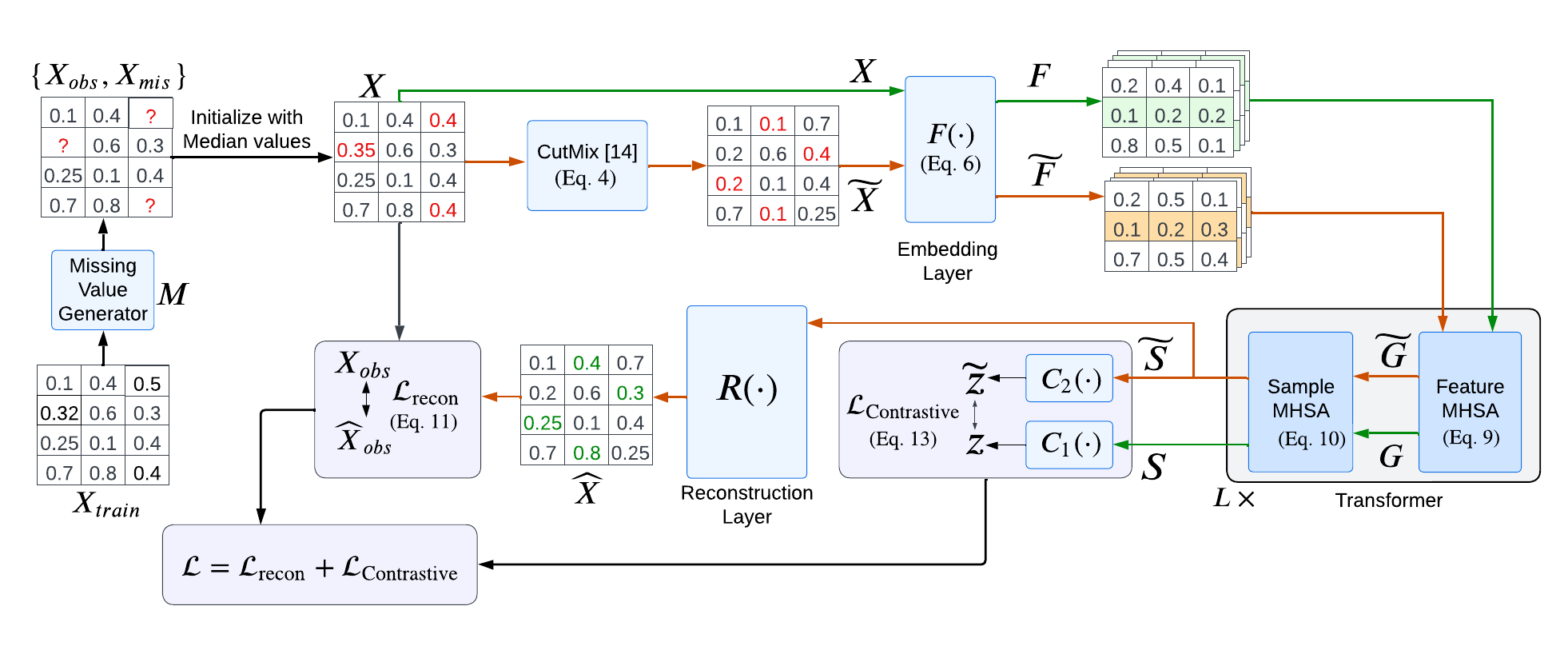}
 \vspace{-20pt}
\caption {Proposed missing value imputation framework. Between-feature attention learning is followed by between-sample attention with a contrastive learning framework. MHSA stands for multi-head self-attention.}
\label{fig: figure_method}
\end{figure*}

This section presents the computational steps of the proposed DeepIFSAC method, including the simulation of missing values, CutMix data augmentation, the joint attention-based reconstruction of missing values, and the contrastive learning framework. These steps are presented in Figure~\ref{fig: figure_method} and Algorithm 1.

\subsection{Statistical types of missing values}
\label{subsec:missing_generation}

A data matrix X with observed values ($X_{obs}$) and missing ($X_{mis}$) values can use a binary matrix $M$ of the same size to identify the position of missing values, where 1 corresponds to observed and 0 corresponds to missing values. The binary matrix can be simulated for one of the three types of statistical missing values: missing completely at random (MCAR), missing at random (MAR), and missing not at random (MNAR). First, MCAR occurs when the probability that a value $M^{i}_{j}$ is missing is independent of both observed $X_{obs}$ and missing data $X_{mis}$, as shown in Equation \ref{eq:mcar}.
\begin{equation}\label{eq:mcar}
    P (M = 0) = P (M = 0~|~X_{\text{obs}}, X_{\text{mis}})
\end{equation}
MCAR happens when someone forgets to enter data or data collection is interrupted due to sensor failure. Second, the MAR type occurs when missing values of a variable depend on the observed values of another variable in Equation \ref{eq:mar}.
\begin{equation}\label{eq:mar}
    P (M = 0~|~X_{\text{obs}}) = P (M = 0~|~X_{\text{obs}}, X_{\text{mis}})
\end{equation}
For example, certain medical lab values can be missing or not recommended by a physician because the patient's age is below eighteen. Third, MNAR occurs when the missing probability depends on the missing values in Equation \ref{eq:mnar}.
\begin{equation}\label{eq:mnar}
    P (M = 0~|~X_{\text{mis}}) = P (M = 0~|~X_{\text{obs}}, X_{\text{mis}})
\end{equation}
For example, the body weight of an obese person may not be recorded during a survey because of personal preference. Therefore, the weight value can be missing because the value is too high. We hypothesize that the same imputation method may not be the best choice for all missing value types, which necessitates benchmarking across different missing value types.

\subsection{Data augmentation}\label{subsec: datamanipulation}

The data augmentation step uses a mask matrix (I) to alter randomly selected values during mini-batch training. The masking and filling steps in data augmentation aim to improve the uncertainty of missing value estimations and enhance model robustness to varying missing patterns. Recent MVI studies perform data augmentation by replacing randomly selected values with zeros~\citep{lee2023, Midaspy_Lall_2022}. Instead of using random values, noise, and zeros for replacement, CutMix~\citep{cutmix_yun2019} uses actual values from other observed samples for replacement to improve data augmentation. The unique masking and filling strategy of CutMix has improved the deep representation of tabular data in classification tasks~\citep{rabbani2024, SAINT_Somepalli2022, scarf}, which we adopt for missing value imputation tasks. The CutMix method replaces a feature value of one sample ($x^i$) with the same feature value of another randomly selected sample ($x^j$), as shown in Equation \ref{eq:cutmix}. 

\begin{equation}\label{eq:cutmix}
    \Tilde{X}^i = X^i \odot I^i + X^j \odot (1 - I^i) 
\end{equation}
The corrupted samples ($\Tilde{X}$) and their uncorrupted versions (X) are taken to subsequent attention learning in Figure~\ref{fig: figure_method}.

\subsection{Missing value estimation}
The transformer architecture \citep{Vaswani2017}, with residual learning \citep{ResNet_he2016} and dots-scaled attention, has been tailored for tabular data classification in recent studies~\citep{SAINT_Somepalli2022, NPT_Kossen2021, FTT_Gorishniy2021}. These state-of-the-art transformer architectures inspire the proposed imputation of tabular data. While supervised classification maps data (X) to class labels as $X \rightarrow Y$, missing value imputation learns and reconstructs $X_{missing}$ from $X_{observed}$ without involving class labels. Therefore, the impact of class (Y) distribution imbalance is not considered in data imputation. Without involving class labels, the proposed reconstruction of missing values through attention learning has the following steps.

\begin{algorithm}
\caption{Proposed DeepIFSAC Algorithm}
\label{alg:deepifsa_algo}
\begin{algorithmic}
\STATE \textbf{Input:} Data $X =\{X_{obs}, X_{mis}\}$ with missing values
\STATE \textbf{Output:} Imputed data matrix, $X = \{X_{obs}, X_{imputed}\}$
\FOR {$t = 1 \rightarrow n\_epochs$}
    \STATE Corrupted data via CutMix, $\Tilde{X}$ = \{X, I\}, Equation~\ref{eq:cutmix}
    \STATE Feature Embeddings, \{$F$, $\Tilde{F}$\} $\leftarrow$ \{X, $\Tilde{X}$\}, Equation~\ref{eq:feature_embeddings},\ref{eq:final_feature_embeddings} 
        \STATE $\{Q_h, K_h, V_h\} \leftarrow F$, $\{\Tilde{Q}_h, \Tilde{K}_h, \Tilde{{V}_h}\}\leftarrow  \Tilde{F}$, Equation~\ref{eq:QKV}
         \STATE Attention scores for each head $h$:
        \STATE $\{head_h, \Tilde{head}_h\} \leftarrow
        \{ \{ Q_h, K_h, V_h\}, \{ \Tilde{Q}_h, \Tilde{K}_h, \Tilde{{V}_h}\}  \}$, Equation~\ref{eq:self-attention}
        \STATE \textbf{Between-feature attention}: 
        \STATE \{G, $\Tilde{G}$\} $\leftarrow$ MHSA(\{$F$, $\Tilde{F}$\}), Equation~\ref{eq:MHSA}
        \STATE \textbf{Between-sample attention}: 
        \STATE $G'$ $\leftarrow$ Reshape G $\in \mathbb{R}^{b \times n, d}$, MHSA between-samples 
        \STATE $S' \leftarrow \text{MHSA}(G')$, Equation~\ref{eq:MHSA_row}
        \STATE $S$ $\leftarrow$ Reshape $S'$ $\in \mathbb{R}^{b, n, d}$ shape for reconstruction
    \STATE Data reconstruction, $\hat{X}$ $\leftarrow$ $\Tilde{S}$
    \STATE $\mathcal{L}_{recon}$ = \{$\hat{X}_{obs}$, $X_{obs}$\}, Equation~\ref{eq:recon_loss}  
    \STATE Projected embedding, \{$Z$, $\Tilde{Z}$\} $\leftarrow$ \{$S$, $\Tilde{S}$\}
    \STATE $\mathcal{L}_{contrastive}$ = \{$Z$, $\Tilde{Z}$\}, Equation~\ref{eq:infoNCE_loss}
    \STATE $\mathcal{L}$ = $\mathcal{L}_{recon}$ + $\mathcal{L}_{contrastive}$
\ENDFOR
\end{algorithmic}
\end{algorithm}


A tabular data set with \( b \) samples and \( n \) features is transformed into a new feature space. Each feature \( j \) is transformed into a \( d \)-dimensional embedding using a linear layer characterized by learnable weight parameters \( W_j \) and bias parameters \( B_j \), as follows.   
\begin{equation}\label{eq:feature_embeddings}
{f}_{j} = W_j {x}_{j} + B_{j} \quad \text{for } j \in \{1, 2, ..., n\}
\end{equation}
Subsequently, \( d \)-dimensional embeddings of individual features are stacked to create a feature embedding tensor \( F \) of size \((b, n, d)\), as in Equation \ref{eq:final_feature_embeddings}.
\begin{align}\label{eq:final_feature_embeddings}
    {F(X)} = \text{stack}~[ f_1, f_2, ..., f_n ]
\end{align}
Two versions of the feature embedding tensor are obtained, one for the uncorrupted input $X$ as \( F \) and the other for the CutMix corrupted samples $\Tilde{X}$ as $\Tilde{F}$, as shown in Figure~\ref{fig: figure_method}.

\subsubsection{Attention between features}

The feature embedding \( F \) is transformed into a feature space \( G \), weighted by the attention between features using the dot-scaled attention mechanism~\citep{Vaswani2017}. Specifically, the embedding \( F \) used to obtain queries (\(Q_h\)), keys (\(K_h\)), and values (\(V_h\)) for the $h$-th attention head is shown in Equation~\ref{eq:QKV}. 
\begin{equation}\label{eq:QKV}
Q_h = F W_h^Q, \quad K_h = F W_h^K, \quad V_h = F W_h^V
\end{equation}
Here, $W^Q_h$, $W^K_h$, and $W^V_h$ are the learnable weight parameters with the shape of $(d, d_H)$, where $d = H*d_H$ for a total of $H$ self-attention heads. The term \( Q_h\) represents the transformed representation of a specific feature within a data point. \( K_h \) encodes all features as keys for comparison with \( Q_h \), determining the relevance between features. Attention scores are obtained from the dot product $Q_{h}K_{h}^T$, which captures the dependency between features within the same data point. Next, \( V_h \) stores the feature representation to be weighted by the attention scores for the output of an attention head. In a dot-scaled attention framework, each attention head results in an embedding ($head_h$) weighted by the attention between features in Equation~\ref{eq:self-attention}.
\begin{equation}\label{eq:self-attention}
head_{h} = \text{softmax}(\frac{Q_{h} (K_{h})^T} {\sqrt{d_H}}) V_{h}
\end{equation}
The output of each attention head has $(b, n, d_H)$ dimensions. In multi-head self-attention (MHSA), the outputs of \( H \) self-attention heads are concatenated to create a tensor of shape $(b, n, d)$.  This tensor is then projected onto a learnable weight matrix $W^C$ with shape $(d, d)$ to obtain the final attention-weighted embedding ($G$) in Equation \ref{eq:MHSA}.
\begin{equation}\label{eq:MHSA}
G = \text{MHSA}(F) =  \text{concat}(\text{head}_1, ..., \text{head}_H) W^C, \quad G \in \mathbb{R}^{(b, n, d)}
\end{equation}
Likewise, the other view of G, obtained by embedding $\Tilde{F}$, is denoted $\Tilde{G}$. 
\subsubsection{Attention between samples}

 The attention between samples is applied to the attention between features embedding (G) obtained in the previous section. First, $G$ is reshaped to $G'$ with dimensions $(1, b, n \times d)$ to ensure that attention operates across b samples instead of features while preserving $n$ feature representations. MHSA (Equation~\ref{eq:MHSA}) operates on $G'$ to produce an embedding $S'$ weighted by between sample attention. $S'$ embedding is then reshaped to $(b, n, d)$ to restore the original embedding structure in Equation~\ref{eq:MHSA_row}, which now incorporates both the attention between features and between samples.

\begin{equation} \label{eq:MHSA_row}
\begin{aligned}
    &\begin{aligned}
        G' & \leftarrow \text{reshape}(G), \quad G' \in \mathbb{R}^{(1, b, n \times d)} \\
        S' &= \text{MHSA}(G'), \quad S' \in \mathbb{R}^{(1, b, n \times d)} \\
        S & \leftarrow \text{reshape}(S'), \quad S \in \mathbb{R}^{(b, n, d)}
    \end{aligned}
    \quad \Bigg\} \vphantom{\begin{aligned} G' \\ S' \\ S \end{aligned}}
\end{aligned}
\end{equation}

The embedding of between-feature attention $\Tilde{G}$ of the CutMix corrupted input undergoes the same transformation of Equation~\ref{eq:MHSA_row} to incorporate attention between samples and result in joint attention embedding $\Tilde{S}$.

\subsubsection{Data reconstruction from attention embedding}\label{subsec:Feat_recon}
A stack of multilayer perceptron (MLP) heads $R(\cdot)$ in Figure~\ref{fig: figure_method} reconstructs the original data (X) from the joint attention embedding ($\Tilde{S}$). When MVI is obtained from attention between features only, $X$ is reconstructed from its corresponding embedding ($\Tilde{G}$). MVI using only the attention between samples is obtained as follows. The embedding F is reshaped to $(1, b, n \times d)$ and then weighted by the attention between the samples using MHSA (Equation~\ref{eq:MHSA}). Next, the embedding with attention between samples is used to reconstruct $X$ for MVI. It is noteworthy that the input ($X$) is reconstructed from the corrupted embeddings $\Tilde{S}$ or ($\Tilde{G}$) instead of its uncorrupted versions. Data reconstruction from the embedding of noisy inputs, similar to a denoising autoencoder, improves the robustness of the estimation of missing values. The process of imputing missing values can introduce noise or data corruption.  The augmented embeddings, generated by CutMix~\citep{cutmix_yun2019}, introduce variations that prevent the model from memorizing missing value patterns. Reconstruction of $X$ from uncorrupted embeddings would limit generalization to unseen missing patterns and make MVI more susceptible to data corruption or noise.

The proposed MVI model is trained using a data reconstruction loss ($\mathcal{L}_{recon}$), which is the mean squared error ($\mathcal{L}_{num}$) between the observed values of the original ($X^{obs}_j$) and their reconstructed ($\hat{X}^{obs}_j$) counterparts, as follows.
\begin{equation}\label{eq:recon_loss_num}
\mathcal{L}_{num} (\hat{X}^{obs}_j, X^{obs}_j) = \frac{1}{b} \sum_{i=1}^{b} M^i_j \left(x^i_j - \hat{x}^i_j\right)^2
\end{equation}

\begin{equation}\label{eq:recon_loss}
\mathcal{L}_{recon} = \sum_{j=1}^{n} \mathcal{L}_{num}\bigl(\hat{X}^{obs}_j, X^{obs}_j\bigr)
\end{equation}

Here, $X^{obs}_j$ and $\hat{X}^{obs}_j$ are the original and reconstructed values of the $j$-th feature, respectively. $\mathcal{L}_{recon}$ is the sum of the reconstruction loss for all features. Reconstruction of missing values ($X_{missing}$) is learned from observed ground truth values ($X_{observed}$) following $X_{missing} \leftarrow X_{observed}$. 
\subsection{Contrastive representation learning}\label{subsec: Contrastive}
 
DeepIFSAC extends the proposed joint feature and sample attention with a contrastive learning framework. Contrastive learning results in an embedding where similar samples 
(\(Z_i\) and \(\Tilde{Z}_i\)) are pulled closer and dissimilar ones (\(Z_i\) and \(Z_k, i \neq k\)) are pushed apart \citep{infoNCE1}. As shown in Figure~\ref{fig: figure_method}, the new embeddings $Z$ and $\widetilde{Z}$ are obtained using two MLP subnetworks, $z^i = C_1(S^i)$ and $\Tilde{z}^i = C_2(\widetilde{S}^i)$ from the corresponding attention embeddings, $S$ and $\widetilde{S}$. The infoNCE loss using $S$ and $\widetilde{S}$ is optimized in Equation~\ref{eq:infoNCE_loss}.   
\begin{eqnarray}\label{eq:infoNCE_loss}
    \mathcal{L}_{Contrastive} = - \sum_{i=1}^b log \frac{\exp~(sim~(z^i, \Tilde{z}^i) / \tau)}{\sum\limits_{k=1}^{b}\exp~(sim~(z^i, \Tilde{z}^k) / \tau)}
\end{eqnarray}
 Here, $sim(.)$ represents the cosine similarity between two embeddings. $b$ is the batch size. The proposed contrastive learning loss (Equation~\ref{eq:infoNCE_loss}) and the data reconstruction loss from attention embeddings (Equation~\ref{eq:recon_loss}) are jointly optimized, presented as $\mathcal{L}$ in Algorithm 1.


\section{Experiments}
This section presents the experimental setup, including the selection of tabular data sets, the system environment, baseline methods, model training, and evaluation. Three missing value types (MCAR, MNAR, and MAR) are simulated for the missing rates 10\%, 30\%, 50\%, 70\%, and 90\%. The real patient data from the All of Us EHR have 35\% values missing. 

\begin{table}[t]
\caption{Tabular data sets used in this study, identified by OpenML ID numbers. Electronic health records (EHR) data sets of heart failure patients are obtained from All of Us~\citep{allofus}. The difficulty is based on the contrast between linear and non-linear classifier performances~\citep{Grinsztajn2022} 
}
\scalebox{0.75}{
\begin{tabular}{llcccc}
\toprule
{ID} & Data set &  Samples &  Features &  Classes &  Difficulty \\ \midrule
11 & Balance-scale & 625 & 4 & 3 & Easy\\
37 & Diabetes & 768 & 8 & 2 & Easy\\
54 & Vehicle & 846 & 18 & 4 & Easy\\
187 & Wine & 178 & 13 & 3 & Easy\\
458 & Analcatdata & 841 & 70 & 4 & Easy\\
1464 & Blood-transfusion & 748 & 5 & 2 & Easy\\
\midrule
1049 & pc4 & 1458 & 37 & 2 & Hard\\
1050 & pc3 & 1563 & 37 & 2 & Hard \\
1067 & Kc1 & 2109 & 21 & 2 & Hard \\
1068 & Pc1 & 1109 & 21 & 2 & Hard \\
1497 & Wall-robot-navigation & 5456 & 24 & 4 & Hard \\
40982 & Steel-plates-fault & 1941 & 27 & 7 & Hard \\
\midrule 
EHR & Heart failure & 2000 & 54 & 2 & - \\
\bottomrule
\end{tabular}
}
\label{table_dataset}
\vspace{-10pt}
\end{table}

\subsection{Tabular data set selection} \label{subsec:tab-dataset}
Table \ref{table_dataset} presents the tabular data sets used in this article. Unlike image data, there is no best learning algorithm due to the heterogeneity of tabular data sets~\citep{rabbani2024}. Therefore, twelve different data sets are selected from the OpenML~\citep{OpenML2013} repository to ensure the robustness of the imputation methods.  The sample size of these data sets ranges from 178 to 5456, with the feature dimension between 4 and 70. Tabular data sets are grouped equally into hard and easy types according to the performance gap between linear and non-linear classifiers~\citep{Grinsztajn2022}.  Notably, most tabular data sets used in the literature and those available in practice have much smaller sample sizes than image and text data sets. Although data sets with large sample sizes better optimize deep learning, data sets with standard sample sizes ensure practical relevance and avoid selection bias. Furthermore, we evaluate the quality of imputed medical data in a heart failure classification task to demonstrate the practical adoption of imputation methods.

\subsection{System environment}
All experiments related to public tabular data sets are carried out on an Ubuntu 22.04 machine that features an Intel (R) Xeon (R) W-2265 CPU @ 3.70GHz, 64GB RAM, and a Quadro RTX A4000 16GB GPU. The processor has 24 logical cores. We use deep learning methods and modules from the PyTorch package, which automatically uses multiple CPUs for parallel processing. All experiments involving All of Us health records data must be conducted within a secure cloud workbench to comply with data sharing and privacy policy. We maintain a uniform system environment by provisioning cloud instances with specifications similar to those of our Ubuntu machine.

\subsection{Baseline methods}
The imputation performance of the proposed attention network is compared to four major categories of state-of-the-art imputation solutions: 1) single value imputation, 2) statistical methods, 3) machine learning methods and 4) state-of-the-art deep learning methods, including GAN, diffusion, and attention-based methods. The single value imputation, such as the median value imputation, is recommended when the missing rate is low~\citep{Mera2021}. MICE~\citep{mice} is the most recognized and used model in statistics and health science. A competitive machine learning model, missForest~\citep{missForest}, uses one of the most successful models for tabular data (random forest) to impute missing values. KNN has shown competitive performance in imputation tasks~\citep{Liao2014}. Recent deep imputation baselines include self-attention based DSAN~\citep{lee2023}, denoising autoencoder based MIDASpy~\citep{Midaspy_Lall_2022}, diffusion-based Diffputer~\citep{Diffputer}, and the GAN-based generative model, GAIN\citep{yoon2018gain}. Several deep imputation methods, including AimNet~\citep{aimnet} and Missdiff\citep{ouyang2023missdiff}, are excluded from the baselines because the relevant source code is not publicly available.

\begin{figure*}[t]
\centering
 \includegraphics[trim=0cm 0.5cm 0cm 0cm, width=1.0\textwidth] {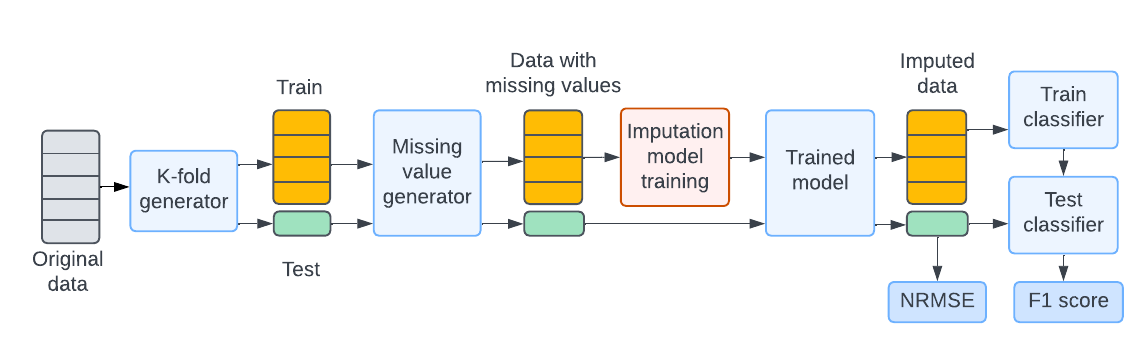}
\caption { 
Data splits for evaluating imputation model performance and quality of imputed data in classification tasks.}
\label{fig: method_pipeline}
\end{figure*}

\subsection{Model setup}

All deep learning models are trained with a batch size of 128 samples using the Adam optimizer after setting $\beta_1 = 0.9$, $\beta_2 = 0.999$, a decay rate of 0.01, and a learning rate of 0.0001. Attention-based models have an embedding size d of 32, eight attention heads (H), and six sequential MHSA blocks (L). The same hyperparameter setting is used in state-of-the-art attention-based models for tabular data~\citep{SAINT_Somepalli2022}. In addition, we apply a dropout rate of 0.1. We train each model for 1000 epochs on training data samples and use the trained model to reconstruct and impute missing values on the left-out test data fold. 

\subsection{Model evaluation} \label{subsec:model-eval}
Figure \ref{fig: method_pipeline} shows the pipeline to evaluate the performance of the imputation models. We split a data set into five folds to train the imputation model with four data folds and test the trained model to impute the left-out data fold. The average imputation error is obtained across five left-out test folds. A fixed random seed ensures the same train-test splits in all experiments. As typical in practice, the data standardization obtained for training data folds is applied to training and test folds. The same fold separation is considered to initialize missing values using the median values. The difference between the model-imputed and ground truth values for the test data fold is measured using the normalized root mean squared error (NRMSE), as shown in Equation \ref{eq:nrmse1}. 
\begin{equation}\label{eq:nrmse1}
NRMSE = \frac{1}{N} \sum_{j=1}^N \sqrt{\frac{\mathbb{E}[(X_j - \hat{X_j})^2]}{\text{Var}(X_j)}}
\end{equation}
Here, N is the number of features. Var ($X_j$) is the variance of the j-th feature, and E (.) obtains the mean of squared errors. The RMSE score is normalized for N number of features as the NRMSE score. The NRMSE scores obtained for five test data folds are averaged as the mean NRMSE score to report the imputation error. In contrast, the All of Us EHR data have missing values without ground truths. Therefore, we evaluate the quality of the imputed data in the downstream classification of heart failure patients, similar to~\citep{Kazijevs2023, Samad2022}, using mean F1 scores across the test data folds.

\begin{table*}[t]
\centering
\caption{Average rank order of imputation methods based on NRMSE scores.}
\label{tab:rank_order_table}
\scalebox{0.45}{
\begin{tabular}{lccccccccccccc}
\toprule
\shortstack{Data\\set} 
& \shortstack{\%\\[0.2em]} 
& \shortstack{Median\\[0.2em]} 
& \shortstack{KNN\\[0.2em]} 
& \shortstack{MICE\\[0.2em]\citep{mice}} 
& \shortstack{missForest\\[0.2em]\citep{missForest}} 
& \shortstack{GAIN\\~\citep{yoon2018gain}} 
& \shortstack{Diffputer\\~\citep{diffimpute}} 
& \shortstack{DSAN\\[0.2em]\citep{lee2023}} 
& \shortstack{MIDASpy\\[0.2em]\citep{Midaspy_Lall_2022}} 
& \shortstack{Between-\\feature~\citep{kowsar2024}} 
& \shortstack{Between-\\sample~\citep{kowsar2024}} 
& \shortstack{DeepIFSAC\\(Proposed)} \\
\midrule
 & 10 & 8.75 (2.05) & 4.50 (1.98) & 3.58 (1.31) & 4.33 (2.50) & 8.33 (1.44) & 7.50 (2.35) & 6.92 (1.73) & 8.75 (3.31) & 3.92 (3.29) & 3.83 (3.01) & 3.17 (3.56) \\
 & 30 & 8.33 (2.15) & 4.08 (2.39) & 6.42 (1.98) & 3.83 (2.29) & 7.67 (1.67) & 8.00 (2.13) & 5.17 (1.59) & 10.25 (1.76) & 3.50 (3.00) & 3.67 (3.20) & 3.17 (3.24) \\
MCAR  & 50 & 8.00 (2.30) & 4.50 (1.93) & 5.42 (1.73) & 4.58 (3.00) & 7.17 (1.27) & 8.75 (1.82) & 4.67 (1.67) & 10.67 (0.89) & 3.25 (2.63) & 4.00 (3.49) & 3.50 (3.37) \\
 & 70 & 6.83 (3.21) & 6.92 (3.18) & 6.00 (3.10) & 6.08 (3.26) & 4.83 (2.95) & 6.50 (2.81) & 3.42 (2.68) & 11.00 (0.00) & 5.25 (2.73) & 3.75 (1.76) & 4.33 (1.78) \\
 & 90 & 6.50 (3.15) & 6.25 (3.11) & 6.42 (2.75) & 7.25 (2.93) & 4.33 (2.81) & 5.58 (3.06) & 4.67 (2.93) & 10.92 (0.29) & 6.08 (2.75) & 3.33 (1.97) & 3.50 (1.62) \\
\midrule
 & Avg. & 7.68 (2.57) & 5.25 (2.52) & 5.57 (2.17) & 5.21 (2.80) & 6.47 (2.03) & 7.27 (2.43) & 4.97 (2.12) & 10.32 (1.25) & 4.40 (2.88) & 3.72 (2.69) & {\bf 3.53 (2.71)} \\
\midrule
 & 10 & 8.33 (1.56) & 4.83 (1.99) & 3.75 (1.29) & 4.33 (2.67) & 8.42 (1.62) & 8.08 (2.68) & 6.50 (1.73) & 8.42 (3.73) & 4.33 (3.26) & 3.67 (2.71) & 3.08 (3.42) \\
 & 30 & 8.25 (2.01) & 4.58 (2.81) & 6.75 (1.71) & 4.08 (2.64) & 7.42 (1.78) & 8.25 (2.01) & 5.17 (1.75) & 8.83 (3.93) & 4.00 (2.83) & 4.00 (2.89) & 3.17 (3.33) \\
MAR   & 50 & 8.25 (1.91) & 4.92 (2.11) & 6.33 (2.10) & 4.50 (2.88) & 7.00 (1.60) & 8.83 (1.90) & 4.92 (1.56) & 9.08 (3.87) & 3.50 (2.54) & 4.00 (3.52) & 3.33 (3.34) \\
 & 70 & 7.42 (3.12) & 6.58 (2.64) & 6.08 (2.84) & 6.92 (2.94) & 5.33 (3.26) & 6.33 (3.26) & 2.50 (2.71) & 11.00 (0.00) & 3.25 (1.42) & 4.33 (1.56) & 4.67 (1.83) \\
 & 90 & 6.50 (3.34) & 6.25 (2.60) & 7.33 (2.57) & 8.25 (1.29) & 4.83 (3.33) & 5.83 (2.95) & 3.75 (3.14) & 11.00 (0.00) & 4.08 (1.88) & 3.17 (1.70) & 3.25 (1.96) \\
\midrule
 & Avg. & 7.75 (2.39) & 5.43 (2.43) & 6.05 (2.10) & 5.62 (2.48) & 6.60 (2.32) & 7.46 (2.56) & 4.57 (2.18) & 9.67 (2.31) & 3.83 (2.39) & 3.83 (2.48) & {\bf 3.50 (2.78)} \\
\midrule
 & 10 & 9.50 (1.68) & 4.83 (1.47) & 5.25 (1.91) & 6.50 (1.51) & 7.83 (1.47) & 7.42 (2.31) & 6.33 (3.47) & 5.92 (4.64) & 4.42 (3.06) & 3.08 (2.94) & 2.58 (3.06) \\
 & 30 & 8.83 (1.75) & 6.75 (1.66) & 6.00 (2.13) & 6.17 (1.80) & 6.83 (1.85) & 8.42 (2.61) & 4.83 (2.48) & 6.58 (4.03) & 4.08 (3.15) & 2.75 (2.96) & 2.33 (3.14) \\
MNAR  & 50 & 7.50 (2.32) & 7.67 (1.23) & 5.67 (2.35) & 6.08 (1.24) & 4.92 (2.02) & 7.50 (2.84) & 4.67 (2.10) & 9.92 (2.35) & 3.58 (2.68) & 2.67 (3.68) & 2.50 (3.09) \\
 & 70 & 6.67 (3.98) & 7.00 (2.83) & 7.58 (2.81) & 6.50 (3.15) & 4.83 (3.30) & 5.75 (3.17) & 5.50 (3.42) & 9.08 (2.50) & 4.00 (2.13) & 3.25 (1.76) & 3.25 (1.54) \\
 & 90 & 7.00 (3.72) & 7.17 (3.19) & 7.75 (1.91) & 6.67 (3.63) & 5.00 (2.59) & 4.58 (3.34) & 6.00 (3.38) & 9.00 (2.45) & 4.00 (2.30) & 3.17 (1.53) & 3.08 (1.73) \\
\midrule
 & Avg. & 7.90 (2.69) & 6.68 (2.08) & 6.45 (2.22) & 6.38 (2.27) & 5.88 (2.25) & 6.73 (2.85) & 5.47 (2.97) & 8.10 (3.19) & 4.02 (2.66) & 2.98 (2.57) & {\bf 2.75 (2.51)} \\
\bottomrule
\end{tabular}
}
\end{table*}

\begin{table*}[t]
\centering
\caption{
Mean normalized root mean square error (NRMSE) after imputing MCAR type data at varying missing rates (\%). The bold values highlight the best scores for individual data sets and missing rate pairs.}
\label{tab:imputation_mcar}
\scalebox{0.47}{
\begin{tabular}{lccccccccccccc}
\toprule
\shortstack{Data\\set} & \shortstack{\%\\[0.2em]} & \shortstack{Median\\[0.2em]} & \shortstack{KNN\\[0.2em]} & \shortstack{MICE\\[0.2em]\citep{mice}} & \shortstack{missForest\\[0.2em]\citep{missForest}} & \shortstack{GAIN\\~\citep{yoon2018gain}} & \shortstack{Diffputer\\~\citep{diffimpute}} & \shortstack{DSAN\\[0.2em]\citep{lee2023}} & \shortstack{MIDASpy\\[0.2em]\citep{Midaspy_Lall_2022}} & \shortstack{Between-\\feature~\citep{kowsar2024}} & \shortstack{Between-\\sample~\citep{kowsar2024}} & {\shortstack{DeepIFSAC\\(Proposed)}} \\
\midrule
    & 10 & 0.3 (0.02)   & 0.36 (0.05)  & 0.3 (0.02)   & 0.37 (0.02)  & 0.34 (0.02)  & 0.44 (0.06)  & 0.33 (0.04)  & 0.32 (0.04)  & 0.22 (0.03)  & 0.22 (0.03)  & \bfseries \color{black} 0.22 (0.03) \\
11  & 30 & 0.55 (0.01)  & 0.61 (0.03)  & 0.55 (0.01)  & 0.65 (0.03)  & 0.58 (0.04)  & 0.73 (0.04)  & 0.55 (0.03)  & 0.70 (0.07)  & 0.38 (0.03)  & 0.38 (0.03)  & \bfseries \color{black} 0.38 (0.03) \\
    & 50 & 0.7 (0.02)   & 0.75 (0.04)  & 0.78 (0.07)  & 0.9 (0.08)   & 0.75 (0.03)  & 0.87 (0.02)  & 0.73 (0.02)  & 1.24 (0.10)  & 0.51 (0.03)  & 0.51 (0.02)  & \bfseries \color{black} 0.50 (0.02) \\
\midrule
  & 10 & 0.32 (0.05)  & 0.3 (0.06)   & 0.28 (0.05)  & 0.3 (0.04)   & 0.32 (0.03)  & 0.37 (0.04)  & 0.34 (0.03)  & 0.30 (0.02)  & 0.18 (0.05)  & 0.18 (0.05)  & \bfseries \color{black} 0.17 (0.05) \\
37  & 30 & 0.53 (0.03)  & 0.55 (0.03)  & 0.54 (0.08)  & 0.54 (0.04)  & 0.55 (0.01)  & 0.65 (0.05)  & 0.52 (0.03)  & 0.66 (0.05)  & 0.29 (0.05)  & 0.30 (0.06)  & \bfseries \color{black} 0.29 (0.06) \\
  & 50 & 0.7 (0.01)   & 0.74 (0.02)  & 0.75 (0.03)  & 0.79 (0.04)  & 0.7 (0.02)   & 0.8 (0.03)   & 0.69 (0.03)  & 1.22 (0.14)  & 0.34 (0.05)  & 0.36 (0.06)  & \bfseries \color{black} 0.34 (0.05) \\
\midrule
  & 10 & 0.32 (0.03)  & 0.15 (0.01)  & 0.14 (0.01)  & 0.14 (0.02)  & 0.27 (0.03)  & 0.18 (0.03)  & 0.22 (0.02)  & 0.33 (0.01)  & 0.04 (0.01)  & 0.03 (0.00)  & \bfseries \color{black} 0.02 (0.00) \\
54  & 30 & 0.55 (0.01)  & 0.29 (0.01)  & 0.3 (0.03)   & 0.29 (0.03)  & 0.47 (0.03)  & 0.38 (0.02)  & 0.38 (0.01)  & 0.72 (0.02)  & 0.07 (0.01)  & 0.04 (0.00)  & \bfseries \color{black} 0.04 (0.01) \\
  & 50 & 0.72 (0.01)  & 0.49 (0.02)  & 0.45 (0.02)  & 0.44 (0.03)  & 0.63 (0.02)  & 0.61 (0.03)  & 0.5 (0.02)   & 1.28 (0.03)  & 0.08 (0.01)  & 0.06 (0.01)  & 0.07 (0.01) \\
\midrule
 & 10 & 0.33 (0.05)  & 0.25 (0.05)  & 0.24 (0.05)  & 0.27 (0.05)  & 0.31 (0.03)  & 0.24 (0.05)  & 0.28 (0.02)  & 0.23 (0.04)  & 0.58 (0.06)  & 0.53 (0.06)  & 0.54 (0.05) \\
187 & 30 & 0.54 (0.03)  & 0.43 (0.05)  & 0.54 (0.07)  & 0.45 (0.05)  & 0.51 (0.03)  & 0.47 (0.03)  & 0.44 (0.02)  & 0.51 (0.04)  & 1.04 (0.10)  & 0.97 (0.11)  & 1.04 (0.12) \\
 & 50 & 0.7 (0.04)   & 0.63 (0.03)  & 0.7 (0.07)   & 0.67 (0.06)  & 0.71 (0.03)  & 0.69 (0.03)  & 0.63 (0.02)  & 0.91 (0.05)  & 1.36 (0.13)  & 1.37 (0.16)  & 1.40 (0.17) \\
\midrule
 & 10 & 0.31 (0.01)  & 0.26 (0.01)  & 0.25 (0.01)  & 0.27 (0.01)  & 0.31 (0.01)  & 0.31 (0.01)  & 0.3 (0.01)   & 0.34 (0.01)  & 0.09 (0.01)  & 0.09 (0.01)  & \bfseries \color{black} 0.08 (0.00) \\
458 & 30 & 0.55 (0.01)  & 0.48 (0.01)  & 0.58 (0.03)  & 0.49 (0.01)  & 0.54 (0.01)  & 0.57 (0.01)  & 0.52 (0.00)  & 0.75 (0.02)  & 0.16 (0.00)  & 0.15 (0.03)  & 0.16 (0.00) \\
 & 50 & 0.72 (0.01)  & 0.65 (0.01)  & 0.69 (0.01)  & 0.69 (0.00)  & 0.7 (0.01)   & 0.73 (0.04)  & 0.67 (0.01)  & 1.35 (0.02)  & 0.21 (0.00)  & 0.22 (0.00)  & 0.23 (0.00) \\
 \midrule
& 10 & 0.3 (0.08)   & 0.24 (0.06)  & 0.2 (0.07)   & 0.19 (0.06)  & 0.34 (0.06)  & 0.27 (0.06)  & 0.29 (0.08)  & 0.25 (0.08)  & 0.02 (0.01)  & 0.02 (0.01)  & \bfseries \color{black} 0.02 (0.01) \\
1464& 30 & 0.57 (0.08)  & 0.5 (0.08)   & 0.41 (0.05)  & 0.36 (0.05)  & 0.51 (0.04)  & 4.18 (7.33)  & 0.43 (0.05)  & 0.53 (0.12)  & 0.04 (0.01)  & 0.04 (0.01)  & \bfseries \color{black} 0.03 (0.01) \\
& 50 & 0.76 (0.07)  & 0.67 (0.07)  & 0.62 (0.07)  & 0.68 (0.13)  & 0.68 (0.07)  & 4.43 (8.01)  & 0.61 (0.06)  & 0.93 (0.31)  & 0.06 (0.01)  & 0.05 (0.01)  & \bfseries \color{black} 0.05 (0.01) \\
\midrule
& 10 & 0.33 (0.03)  & 0.18 (0.03)  & 0.18 (0.03)  & 0.14 (0.01)  & 0.25 (0.02)  & 0.25 (0.04)  & 0.23 (0.03)  & 13.30 (6.71) & 0.23 (0.05)  & 0.24 (0.04)  & 0.19 (0.06) \\
1049& 30 & 0.57 (0.03)  & 0.35 (0.03)  & 0.39 (0.08)  & 0.3 (0.03)   & 0.46 (0.03)  & 0.44 (0.03)  & 0.4 (0.02)   & 42.32 (19.37)& 0.38 (0.04)  & 0.38 (0.04)  & 0.39 (0.08) \\
& 50 & 0.75 (0.03)  & 0.53 (0.03)  & 0.54 (0.05)  & 0.45 (0.04)  & 0.62 (0.01)  & 0.69 (0.06)  & 0.53 (0.03)  & 59.11 (25.99)& 0.53 (0.04)  & 0.59 (0.07)  & 0.51 (0.01) \\
\midrule
& 10 & 0.31 (0.03)  & 0.16 (0.03)  & 0.11 (0.01)  & 0.15 (0.03)  & 0.25 (0.01)  & 0.24 (0.13)  & 0.21 (0.02)  & 9.73 (2.26)  & 0.17 (0.02)  & 0.24 (0.07)  & 0.22 (0.08) \\
1050& 30 & 0.56 (0.03)  & 0.33 (0.07)  & 0.38 (0.14)  & 0.3 (0.06)   & 0.47 (0.03)  & 0.46 (0.06)  & 0.4 (0.04)   & 37.46 (33.27)& 0.33 (0.02)  & 0.45 (0.07)  & 0.44 (0.09) \\
& 50 & 0.74 (0.01)  & 0.53 (0.06)  & 0.53 (0.15)  & 0.45 (0.06)  & 0.62 (0.04)  & 0.79 (0.33)  & 0.54 (0.04)  & 56.10 (53.69)& 0.48 (0.03)  & 0.59 (0.06)  & 0.58 (0.08) \\
\midrule
& 10 & 0.31 (0.03)  & 0.12 (0.02)  & 0.13 (0.04)  & 0.13 (0.03)  & 0.25 (0.02)  & 0.22 (0.15)  & 0.15 (0.01)  & 5.12 (1.16)  & 0.10 (0.02)  & 0.12 (0.03)  & \bfseries \color{black} 0.06 (0.02) \\
1067& 30 & 0.57 (0.01)  & 0.24 (0.01)  & 0.32 (0.08)  & 0.24 (0.03)  & 0.45 (0.02)  & 0.4 (0.04)   & 0.3 (0.02)   & 11.15 (3.78) & 0.20 (0.04)  & 0.21 (0.05)  & \bfseries \color{black} 0.16 (0.04) \\
& 50 & 0.74 (0.02)  & 0.44 (0.04)  & 0.41 (0.04)  & 0.34 (0.03)  & 0.59 (0.03)  & 0.73 (0.27)  & 0.41 (0.03)  & 19.95 (6.51) & 0.32 (0.05)  & 0.31 (0.04)  & \bfseries \color{black} 0.31 (0.04)  \\
\midrule
& 10 & 0.32 (0.06)  & 0.18 (0.05)  & 0.14 (0.07)  & 0.19 (0.04)  & 0.29 (0.06)  & 0.19 (0.07)  & 0.23 (0.05)  & 22.01 (18.66)& 0.08 (0.01)  & 0.09 (0.03)  & \bfseries \color{black} 0.04 (0.01) \\
1068& 30 & 0.56 (0.05)  & 0.35 (0.04)  & 0.6 (0.6)    & 0.35 (0.07)  & 0.48 (0.02)  & 0.46 (0.11)  & 0.45 (0.06)  & 95.01 (128.75)& 0.18 (0.04)  & 0.19 (0.06)  & \bfseries \color{black} 0.12 (0.05) \\
& 50 & 0.75 (0.02)  & 0.55 (0.05)  & 0.61 (0.17)  & 0.49 (0.05)  & 0.62 (0.03)  & 0.65 (0.04)  & 0.62 (0.09)  & 25.19 (14.27)& 0.29 (0.04)  & 0.30 (0.03)  & \bfseries \color{black} 0.28 (0.08)  \\
\midrule
& 10 & 0.33 (0.01)  & 0.21 (0.01)  & 0.25 (0.01)  & 0.2 (0.0)    & 0.32 (0.01)  & 0.29 (0.01)  & 0.27 (0.0)   & 0.32 (0.00)  & 0.26 (0.01)  & 0.19 (0.01)  & \bfseries \color{black} 0.18 (0.00) \\
1497& 30 & 0.57 (0.01)  & 0.36 (0.01)  & 0.52 (0.01)  & 0.4 (0.0)    & 0.55 (0.0)   & 0.53 (0.02)  & 0.5 (0.01)   & 0.70 (0.01)  & 0.44 (0.01)  & 0.32 (0.01)  & \bfseries \color{black} 0.32 (0.01) \\
& 50 & 0.75 (0.0)   & 0.52 (0.01)  & 0.65 (0.01)  & 0.59 (0.01)  & 0.72 (0.02)  & 0.75 (0.01)  & 0.65 (0.01)  & 1.25 (0.01)  & 0.56 (0.02)  & 0.44 (0.00)  & \bfseries \color{black} 0.43 (0.01) \\
\midrule
& 10 & 0.33 (0.0)   & 0.16 (0.02)  & 0.17 (0.02)  & 0.13 (0.01)  & 0.28 (0.01)  & 0.33 (0.34)  & 0.25 (0.01)  & 239.65 (85.17)& 0.45 (0.09)  & 0.42 (0.10)  & 0.46 (0.25) \\
40982& 30 & 0.56 (0.01)  & 0.31 (0.01)  & 0.55 (0.49)  & 0.27 (0.01)  & 0.49 (0.01)  & 0.54 (0.30)  & 0.44 (0.04)  & 833.29 (209.93)& 0.72 (0.04)  & 0.78 (0.18)  & 0.65 (0.04) \\
& 50 & 0.72 (0.01)  & 0.55 (0.01)  & 0.5 (0.02)   & 0.45 (0.01)  & 0.65 (0.01)  & 0.73 (0.14)  & 0.6 (0.1)    & 1807.24 (392.47)& 1.07 (0.04)  & 1.66 (0.07)  & 1.13 (0.11) \\
\bottomrule
\end{tabular}
}
\vspace{-10pt}
\end{table*}

\begin{table*}[t]
\centering
\caption{Mean normalized root mean square error (NRMSE) after imputing MAR type data at varying missing rates (\%). The bold values highlight the best scores for individual data sets and missing rate pairs.}
\label{tab:imputation_mar}
\scalebox{0.47}{
\begin{tabular}{lccccccccccccc}
\toprule
\shortstack{Data\\set} & \shortstack{\%\\[0.2em]} & \shortstack{Median\\[0.2em]} & \shortstack{KNN\\[0.2em]} & \shortstack{MICE\\[0.2em]\citep{mice}} & \shortstack{missForest\\[0.2em]\citep{missForest}} & \shortstack{GAIN\\~\citep{yoon2018gain}} & \shortstack{Diffputer\\~\citep{diffimpute}} & \shortstack{DSAN\\[0.2em]\citep{lee2023}} & \shortstack{MIDASpy\\[0.2em]\citep{Midaspy_Lall_2022}} & \shortstack{Between-\\feature~\citep{kowsar2024}} & \shortstack{Between-\\sample~\citep{kowsar2024}} & {\shortstack{DeepIFSAC\\(Proposed)}} \\
\midrule
    & 10 & 0.31 (0.01) & 0.36 (0.03) & 0.31 (0.01) & 0.40 (0.02) & 0.35 (0.02) & 0.44 (0.04) & 0.33 (0.02) & 0.32 (0.01) & 0.23 (0.01) & 0.23 (0.01) & \bfseries \color{black} 0.22 (0.00) \\
11    & 30 & 0.55 (0.01) & 0.63 (0.01) & 0.55 (0.01) & 0.66 (0.04) & 0.60 (0.03) & 0.69 (0.03) & 0.56 (0.02) & 0.69 (0.01) & 0.39 (0.01) & 0.40 (0.01) & \bfseries \color{black} 0.38 (0.01) \\
    & 50 & 0.71 (0.00) & 0.79 (0.01) & 0.81 (0.09) & 0.87 (0.04) & 0.77 (0.03) & 0.87 (0.03) & 0.73 (0.01) & 1.24 (0.01) & 0.51 (0.01) & 0.51 (0.01) & \bfseries \color{black} 0.51 (0.02) \\
\midrule
    & 10 & 0.30 (0.02) & 0.29 (0.03) & 0.26 (0.02) & 0.28 (0.02) & 0.34 (0.05) & 0.41 (0.06) & 0.32 (0.03) & 0.30 (0.01) & 0.13 (0.02) & 0.18 (0.06) & 0.14 (0.04) \\
37    & 30 & 0.57 (0.04) & 0.58 (0.03) & 0.57 (0.03) & 0.57 (0.03) & 0.56 (0.02) & 0.65 (0.04) & 0.53 (0.04) & 0.65 (0.03) & 0.29 (0.03) & 0.27 (0.03) & \bfseries \color{black} 0.25 (0.02) \\
    & 50 & 0.73 (0.03) & 0.77 (0.03) & 0.79 (0.04) & 0.77 (0.03) & 0.70 (0.02) & 0.79 (0.04) & 0.70 (0.04) & 1.14 (0.10) & 0.36 (0.04) & 0.33 (0.05) & 0.36 (0.06) \\
\midrule
    & 10 & 0.32 (0.02) & 0.14 (0.01) & 0.13 (0.01) & 0.14 (0.02) & 0.32 (0.01) & 0.20 (0.02) & 0.22 (0.02) & 0.33 (0.01) & 0.04 (0.01) & 0.02 (0.00) & 0.03 (0.00) \\
54    & 30 & 0.55 (0.01) & 0.29 (0.03) & 0.32 (0.03) & 0.26 (0.01) & 0.51 (0.02) & 0.47 (0.13) & 0.37 (0.01) & 0.73 (0.01) & 0.07 (0.01) & 0.04 (0.01) & 0.05 (0.01) \\
    & 50 & 0.72 (0.02) & 0.51 (0.02) & 0.46 (0.07) & 0.41 (0.02) & 0.65 (0.02) & 0.64 (0.05) & 0.53 (0.02) & 1.24 (0.03) & 0.09 (0.01) & 0.06 (0.01) & \bfseries \color{black} 0.06 (0.01) \\
\midrule
   & 10 & 0.32 (0.03) & 0.25 (0.04) & 0.24 (0.03) & 0.29 (0.03) & 0.27 (0.04) & 0.23 (0.03) & 0.25 (0.04) & 0.25 (0.01) & 0.58 (0.08) & 0.48 (0.06) & 0.57 (0.07) \\
187   & 30 & 0.55 (0.03) & 0.45 (0.03) & 0.68 (0.12) & 0.46 (0.04) & 0.50 (0.04) & 0.49 (0.05) & 0.45 (0.04) & 0.53 (0.02) & 0.99 (0.08) & 0.94 (0.07) & 1.00 (0.11) \\
   & 50 & 0.70 (0.02) & 0.63 (0.04) & 0.70 (0.10) & 0.65 (0.03) & 0.69 (0.03) & 0.67 (0.03) & 0.62 (0.03) & 0.91 (0.04) & 1.27 (0.16) & 1.35 (0.16) & 1.43 (0.20) \\
\midrule
   & 10 & 0.32 (0.01) & 0.27 (0.01) & 0.26 (0.01) & 0.27 (0.01) & 0.31 (0.01) & 0.32 (0.01) & 0.31 (0.01) & 0.34 (0.01) & 0.09 (0.01) & 0.09 (0.01) & 0.09 (0.00)\\
458   & 30 & 0.56 (0.00) & 0.49 (0.00) & 0.58 (0.03) & 0.49 (0.01) & 0.55 (0.01) & 0.59 (0.03) & 0.52 (0.01) & 0.75 (0.01) & 0.16 (0.01) & 0.17 (0.00) & \bfseries \color{black} 0.16 (0.00) \\
   & 50 & 0.72 (0.01) & 0.65 (0.01) & 0.69 (0.01) & 0.69 (0.01) & 0.70 (0.01) & 0.73 (0.03) & 0.67 (0.01) & 1.34 (0.02) & 0.21 (0.00) & 0.22 (0.00) & 0.23 (0.01) \\
\midrule
  & 10 & 0.36 (0.08) & 0.27 (0.05) & 0.20 (0.03) & 0.24 (0.06) & 0.34 (0.09) & 0.35 (0.11) & 0.29 (0.03) & 0.24 (0.03) & 0.02 (0.00) & 0.02 (0.00) & \bfseries \color{black} 0.02 (0.00) \\
1464  & 30 & 0.56 (0.03) & 0.47 (0.03) & 0.38 (0.02) & 0.41 (0.02) & 0.56 (0.05) & 2.72 (4.86) & 0.40 (0.02) & 0.45 (0.05) & 0.04 (0.01) & 0.04 (0.00) & 0.04 (0.01) \\
  & 50 & 0.70 (0.04) & 0.60 (0.04) & 0.55 (0.04) & 0.63 (0.09) & 0.68 (0.03) & 1.67 (1.98) & 0.59 (0.05) & 0.85 (0.12) & 0.06 (0.01) & 0.06 (0.01) & \bfseries \color{black} 0.05 (0.01) \\
\midrule
  & 10 & 0.33 (0.02) & 0.19 (0.03) & 0.18 (0.03) & 0.16 (0.03) & 0.33 (0.01) & 0.29 (0.06) & 0.22 (0.04) & 0.16 (0.03) & 0.23 (0.04) & 0.22 (0.06) & \bfseries \color{black} 0.16 (0.09) \\
1049  & 30 & 0.56 (0.02) & 0.34 (0.04) & 0.42 (0.15) & 0.3 (0.02) & 0.51 (0.01) & 0.51 (0.06) & 0.39 (0.03) & 41.43 (16.93) & 0.39 (0.03) & 0.37 (0.05) & 0.34 (0.02) \\
  & 50 & 0.74 (0.02) & 0.53 (0.04) & 0.57 (0.06) & 0.45 (0.04) & 0.63 (0.01) & 0.87 (0.44) & 0.58 (0.08) & 58.39 (10.34) & 0.54 (0.04) & 0.66 (0.04) & 0.52 (0.05) \\
\midrule
  & 10 & 0.33 (0.02) & 0.18 (0.02) & 0.16 (0.04) & 0.16 (0.04) & 0.34 (0.04) & 0.34 (0.17) & 0.25 (0.06) & 16.78 (11.53) & 0.20 (0.03) & 0.25 (0.06) & 0.24 (0.05) \\
1050  & 30 & 0.57 (0.04) & 0.34 (0.02) & 0.38 (0.12) & 0.29 (0.04) & 0.54 (0.03) & 0.51 (0.02) & 0.40 (0.04) & 49.64 (37.46) & 0.37 (0.04) & 0.46 (0.08) & 0.46 (0.06) \\
  & 50 & 0.76 (0.03) & 0.54 (0.06) & 0.74 (0.55) & 0.45 (0.07) & 0.66 (0.04) & 0.67 (0.02) & 0.56 (0.06) & 73.22 (43.53) & 0.51 (0.03) & 0.62 (0.05) & 0.59 (0.04) \\
\midrule
  & 10 & 0.30 (0.02) & 0.12 (0.03) & 0.12 (0.03) & 0.10 (0.02) & 0.35 (0.06) & 0.29 (0.03) & 0.17 (0.02) & 4.56 (1.26) & 0.11 (0.05) & 0.09 (0.02) & \bfseries \color{black} 0.06 (0.02) \\
1067  & 30 & 0.59 (0.01) & 0.27 (0.01) & 0.52 (0.49) & 0.26 (0.02) & 0.51 (0.04) & 0.49 (0.05) & 0.33 (0.01) & 14.38 (7.14) & 0.22 (0.03) & 0.22 (0.02) & \bfseries \color{black} 0.18 (0.05) \\
  & 50 & 0.75 (0.02) & 0.46 (0.04) & 0.43 (0.04) & 0.36 (0.02) & 0.63 (0.01) & 0.67 (0.04) & 0.42 (0.02) & 26.30 (10.97) & 0.35 (0.06) & 0.34 (0.05) & \bfseries \color{black} 0.33 (0.03) \\
\midrule
  & 10 & 0.35 (0.04) & 0.20 (0.07) & 0.18 (0.09) & 0.18 (0.07) & 0.39 (0.06) & 0.39 (0.11) & 0.24 (0.05) & 15.69 (6.88) & 0.10 (0.03) & 0.12 (0.03) & \bfseries \color{black} 0.04 (0.01) \\
1068  & 30 & 0.57 (0.06) & 0.36 (0.07) & 0.56 (0.30) & 0.35 (0.07) & 0.55 (0.06) & 0.53 (0.03) & 0.40 (0.07) & 68.18 (69.63) & 0.20 (0.04) & 0.23 (0.03) & \bfseries \color{black} 0.18 (0.05) \\
  & 50 & 0.75 (0.04) & 0.55 (0.05) & 0.74 (0.25) & 0.51 (0.07) & 0.69 (0.06) & 0.90 (0.43) & 0.52 (0.05) & 42.05 (23.57) & 0.29 (0.07) & 0.29 (0.03) & 0.29 (0.07) \\
\midrule
  & 10 & 0.33 (0.01) & 0.20 (0.01) & 0.26 (0.01) & 0.21 (0.01) & 0.33 (0.00) & 0.29 (0.01) & 0.28 (0.01) & 0.32 (0.00) & 0.27 (0.00) & 0.19 (0.01) & \bfseries \color{black} 0.18 (0.00) \\
1497  & 30 & 0.57 (0.01) & 0.35 (0.01) & 0.53 (0.02) & 0.39 (0.01) & 0.57 (0.01) & 0.56 (0.01) & 0.50 (0.01) & 0.70 (0.00) & 0.44 (0.01) & 0.32 (0.00) & \bfseries \color{black} 0.31 (0.01) \\
  & 50 & 0.74 (0.00) & 0.52 (0.01) & 0.65 (0.01) & 0.60 (0.01) & 0.73 (0.00) & 0.75 (0.01) & 0.65 (0.01) & 1.25 (0.01) & 0.56 (0.01) & 0.44 (0.01) & \bfseries \color{black} 0.43 (0.01) \\
\midrule
 & 10 & 0.32 (0.02) & 0.15 (0.01) & 0.20 (0.11) & 0.12 (0.01) & 0.29 (0.01) & 0.31 (0.28) & 0.24 (0.01) & 302.14 (79.73) & 0.45 (0.04) & 0.41 (0.16) & 0.51 (0.29) \\
40982 & 30 & 0.55 (0.01) & 0.31 (0.02) & 0.56 (0.46) & 0.27 (0.01) & 0.49 (0.01) & 0.57 (0.36) & 0.43 (0.01) & 691.86 (146.52) & 0.81 (0.07) & 0.87 (0.28) & 0.61 (0.03) \\
 & 50 & 0.74 (0.02) & 0.56 (0.03) & 0.55 (0.02) & 0.46 (0.03) & 0.67 (0.00) & 0.97 (0.71) & 0.57 (0.02) & 1630.46 (370.94) & 1.12 (0.12) & 1.63 (0.09) & 1.22 (0.19) \\
\bottomrule
\end{tabular}
}
\vspace{-10pt}
\end{table*}

\begin{table*}[t]
\centering
\caption{Mean normalized root mean square error (NRMSE) after imputing MNAR type data at varying missing rates. The bold values highlight the best scores for individual data sets and missing rate pairs.}
\label{tab:imputation_mnar}
\scalebox{0.47}{
\begin{tabular}{lccccccccccccc}
\toprule
\shortstack{Data\\set} & \shortstack{\%\\[0.2em]} & \shortstack{Median\\[0.2em]} & \shortstack{KNN\\[0.2em]} & \shortstack{MICE\\[0.2em]\citep{mice}} & \shortstack{missForest\\[0.2em]\citep{missForest}} & \shortstack{GAIN\\~\citep{yoon2018gain}} & \shortstack{Diffputer\\~\citep{diffimpute}} & \shortstack{DSAN\\[0.2em]\citep{lee2023}} & \shortstack{MIDASpy\\[0.2em]\citep{Midaspy_Lall_2022}} & \shortstack{Between-\\feature~\citep{kowsar2024}} & \shortstack{Between-\\sample~\citep{kowsar2024}} & {\shortstack{DeepIFSAC\\(Proposed)}} \\
\midrule
    & 10 & 0.45 (0.0)  & 0.44 (0.02) & 0.46 (0.0)  & 0.46 (0.04) & 0.47 (0.01) & 0.47 (0.02) & 0.49 (0.06) & 0.29 (0.01) & 0.32 (0.01) & 0.32 (0.01) & 0.31 (0.00) \\
11    & 30 & 0.77 (0.01) & 0.8 (0.02)  & 0.81 (0.01) & 0.8 (0.02)  & 0.82 (0.02) & 0.82 (0.02) & 0.78 (0.02) & 0.70 (0.01) & 0.55 (0.01) & 0.54 (0.01) & \bfseries \color{black} 0.54 (0.01) \\
    & 50 & 0.78 (0.02) & 0.9 (0.03)  & 0.9 (0.02)  & 0.81 (0.05) & 0.9 (0.06)  & 0.89 (0.04) & 0.83 (0.02) & 1.13 (0.01) & 0.62 (0.02) & 0.58 (0.02) & 0.60 (0.02) \\
\midrule
    & 10 & 0.62 (0.02) & 0.6 (0.03)  & 0.58 (0.02) & 0.59 (0.02) & 0.61 (0.02) & 0.62 (0.02) & 0.63 (0.01) & 0.29 (0.01) & 0.35 (0.03) & 0.34 (0.03) & 0.34 (0.03) \\
37    & 30 & 0.84 (0.03) & 0.84 (0.02) & 0.8 (0.03)  & 0.83 (0.02) & 0.83 (0.02) & 0.86 (0.02) & 0.82 (0.03) & 0.67 (0.02) & 0.44 (0.04) & 0.43 (0.04) & \bfseries \color{black} 0.43 (0.03) \\
    & 50 & 0.92 (0.01) & 0.93 (0.01) & 0.92 (0.02) & 0.93 (0.02) & 0.91 (0.01) & 0.94 (0.02) & 0.92 (0.01) & 1.11 (0.05) & 0.47 (0.05) & 0.47 (0.05) & \bfseries \color{black} 0.47 (0.05) \\
\midrule
    & 10 & 0.6 (0.03)  & 0.37 (0.08) & 0.35 (0.07) & 0.43 (0.05) & 0.52 (0.04) & 0.41 (0.07) & 0.41 (0.06) & 0.28 (0.01) & 0.07 (0.01) & 0.06 (0.01) & \bfseries \color{black} 0.06 (0.00) \\
54    & 30 & 0.83 (0.01) & 0.69 (0.03) & 0.62 (0.03) & 0.71 (0.05) & 0.79 (0.01) & 0.71 (0.01) & 0.62 (0.03) & 0.60 (0.03) & 0.11 (0.01) & 0.09 (0.01) & \bfseries \color{black} 0.09 (0.01) \\
    & 50 & 0.93 (0.01) & 0.91 (0.01) & 0.83 (0.01) & 0.85 (0.02) & 0.9 (0.01)  & 0.89 (0.01) & 0.78 (0.02) & 1.04 (0.03) & 0.12 (0.01) & 0.11 (0.01) & \bfseries \color{black} 0.11 (0.01) \\
\midrule
   & 10 & 0.6 (0.01)  & 0.47 (0.02) & 0.45 (0.02) & 0.48 (0.02) & 0.49 (0.02) & 0.44 (0.01) & 0.48 (0.01) & 0.25 (0.01) & 1.14 (0.09) & 1.05 (0.10) & 1.08 (0.13) \\
187   & 30 & 0.82 (0.01) & 0.74 (0.02) & 0.75 (0.06) & 0.72 (0.02) & 0.75 (0.02) & 0.72 (0.03) & 0.7 (0.02)  & 0.54 (0.02) & 1.57 (0.11) & 1.51 (0.12) & 1.56 (0.12) \\
   & 50 & 0.93 (0.0)  & 0.93 (0.01) & 0.89 (0.01) & 0.92 (0.02) & 0.89 (0.01) & 0.9 (0.02)  & 0.9 (0.05)  & 0.93 (0.06) & 1.83 (0.12) & 1.87 (0.18) & 1.84 (0.14) \\
\midrule
   & 10 & 0.6 (0.01)  & 0.5 (0.01)  & 0.5 (0.0)   & 0.52 (0.01) & 0.56 (0.01) & 0.53 (0.0)  & 0.53 (0.01) & 0.34 (0.01) & 0.16 (0.00) & 0.15 (0.01) & \bfseries \color{black} 0.15 (0.00) \\
458   & 30 & 0.82 (0.01) & 0.78 (0.01) & 0.75 (0.02) & 0.76 (0.01) & 0.8 (0.01)  & 0.77 (0.01) & 0.81 (0.01) & 0.76 (0.01) & 0.23 (0.00) & 0.23 (0.00) & \bfseries \color{black} 0.23 (0.00) \\
   & 50 & 0.9 (0.0)   & 0.91 (0.01) & 0.92 (0.01) & 0.89 (0.01) & 0.9 (0.01)  & 0.95 (0.08) & 0.95 (0.01) & 1.33 (0.02) & 0.27 (0.00) & 0.26 (0.00) & 0.27 (0.00) \\
\midrule
  & 10 & 0.53 (0.09) & 0.49 (0.07) & 0.5 (0.1)   & 0.51 (0.1)  & 0.5 (0.08)  & 0.5 (0.1)   & 0.52 (0.1)  & 0.21 (0.08) & 0.04 (0.01) & 0.04 (0.01) & \bfseries \color{black} 0.04 (0.01) \\
1464  & 30 & 0.87 (0.01) & 0.85 (0.03) & 0.85 (0.01) & 0.88 (0.02) & 0.83 (0.02) & 1.41 (1.22) & 0.86 (0.01) & 0.35 (0.05) & 0.07 (0.01) & 0.07 (0.01) & \bfseries \color{black} 0.07 (0.01) \\
  & 50 & 0.91 (0.04) & 0.89 (0.04) & 0.91 (0.04) & 0.91 (0.05) & 0.88 (0.03) & 0.93 (0.04) & 0.9 (0.05)  & 0.71 (0.28) & 0.08 (0.01) & 0.08 (0.01) & \bfseries \color{black} 0.08 (0.01) \\
\midrule
  & 10 & 0.64 (0.04) & 0.49 (0.05) & 0.51 (0.04) & 0.52 (0.06) & 0.55 (0.07) & 0.56 (0.17) & 0.47 (0.06) & 5.90 (2.50) & 0.48 (0.03) & 0.49 (0.02) & \bfseries \color{black} 0.45 (0.11) \\
1049  & 30 & 0.86 (0.01) & 0.81 (0.02) & 0.95 (0.46) & 0.78 (0.01) & 0.81 (0.02) & 1.38 (1.01) & 0.71 (0.03) & 1.46 (0.13) & 0.73 (0.04) & 0.70 (0.02) & \bfseries \color{black} 0.68 (0.03) \\
  & 50 & 0.93 (0.01) & 0.91 (0.01) & 0.87 (0.01) & 0.88 (0.01) & 0.89 (0.02) & 4.44 (7.64) & 0.82 (0.02) & 2.30 (0.32) & 0.84 (0.04) & 0.81 (0.03) & \bfseries \color{black} 0.81 (0.03) \\
\midrule
  & 10 & 0.69 (0.02) & 0.57 (0.05) & 0.55 (0.07) & 0.58 (0.05) & 0.63 (0.04) & 0.59 (0.04) & 0.55 (0.05) & 1.35 (0.28) & 0.54 (0.03) & 0.50 (0.02) & \bfseries \color{black} 0.50 (0.05) \\
1050  & 30 & 0.89 (0.02) & 0.85 (0.03) & 0.8 (0.04)  & 0.82 (0.04) & 0.84 (0.03) & 1.13 (0.59) & 0.74 (0.05) & 1.40 (0.22) & 0.73 (0.03) & 0.70 (0.02) & \bfseries \color{black} 0.64 (0.01) \\
  & 50 & 0.95 (0.01) & 0.94 (0.01) & 0.9 (0.02)  & 0.91 (0.03) & 0.9 (0.02)  & 0.92 (0.02) & 0.85 (0.03) & 1.87 (0.26) & 0.81 (0.04) & 0.78 (0.03) & \bfseries \color{black} 0.78 (0.02) \\
\midrule
  & 10 & 0.75 (0.03) & 0.54 (0.04) & 0.61 (0.03) & 0.59 (0.04) & 0.6 (0.04)  & 0.68 (0.12) & 0.49 (0.03) & 2.07 (0.67) & 0.33 (0.04) & 0.32 (0.05) & \bfseries \color{black} 0.29 (0.07) \\
1067  & 30 & 0.89 (0.02) & 0.77 (0.02) & 0.82 (0.03) & 0.79 (0.03) & 0.76 (0.02) & 0.89 (0.09) & 0.74 (0.04) & 2.23 (0.93) & 0.49 (0.02) & 0.47 (0.02) & \bfseries \color{black} 0.46 (0.02) \\
  & 50 & 0.97 (0.02) & 0.92 (0.02) & 0.94 (0.02) & 0.92 (0.02) & 0.91 (0.02) & 1.02 (0.15) & 0.89 (0.03) & 1.22 (0.14) & 0.58 (0.01) & 0.57 (0.01) & \bfseries \color{black} 0.55 (0.03) \\
\midrule
  & 10 & 0.8 (0.06)  & 0.68 (0.11) & 0.71 (0.07) & 0.71 (0.09) & 0.71 (0.08) & 0.74 (0.09) & 0.65 (0.11) & 1.75 (0.56) & 0.27 (0.06) & 0.26 (0.05) & \bfseries \color{black} 0.26 (0.05) \\
1068  & 30 & 0.95 (0.01) & 0.89 (0.03) & 0.88 (0.03) & 0.88 (0.02) & 0.88 (0.02) & 0.96 (0.14) & 0.83 (0.04) & 1.40 (0.42) & 0.44 (0.02) & 0.43 (0.03) & \bfseries \color{black} 0.41 (0.02) \\
  & 50 & 1.0 (0.01)  & 0.98 (0.01) & 0.97 (0.02) & 0.97 (0.02) & 0.94 (0.02) & 0.96 (0.03) & 0.94 (0.03) & 1.50 (0.31) & 0.48 (0.07) & 0.46 (0.06) & 0.46 (0.08) \\
\midrule
  & 10 & 0.72 (0.0)  & 0.59 (0.01) & 0.61 (0.01) & 0.58 (0.01) & 0.62 (0.01) & 0.64 (0.01) & 0.64 (0.01) & 0.32 (0.00) & 0.60 (0.01) & 0.54 (0.01) & 0.51 (0.01) \\
1497  & 30 & 0.95 (0.0)  & 0.87 (0.01) & 0.87 (0.01) & 0.86 (0.01) & 0.86 (0.0)  & 0.87 (0.0)  & 0.84 (0.0)  & 0.72 (0.00) & 0.78 (0.01) & 0.72 (0.00) & \bfseries \color{black} 0.71 (0.00) \\
  & 50 & 1.02 (0.0)  & 1.02 (0.0)  & 0.99 (0.0)  & 0.99 (0.0)  & 0.95 (0.0)  & 0.96 (0.0)  & 0.99 (0.01) & 1.36 (0.00) & 0.84 (0.01) & 0.82 (0.01) & \bfseries \color{black} 0.82 (0.01) \\
\midrule
 & 10 & 0.65 (0.02) & 0.49 (0.04) & 0.49 (0.03) & 0.57 (0.03) & 0.58 (0.02) & 0.55 (0.02) & 0.49 (0.04) & 110.21 (13.21) & 1.10 (0.11) & 1.03 (0.15) & 0.90 (0.08) \\
40982 & 30 & 0.85 (0.0)  & 0.76 (0.01) & 0.72 (0.01) & 0.8 (0.02)  & 0.81 (0.01) & 0.78 (0.01) & 0.69 (0.01) & 277.71 (34.47) & 1.75 (0.05) & 1.73 (0.08) & 1.66 (0.03) \\
 & 50 & 0.94 (0.01) & 0.95 (0.01) & 0.9 (0.01)  & 0.94 (0.01) & 0.92 (0.01) & 0.92 (0.01) & 0.88 (0.01) & 489.48 (44.21) & 2.19 (0.06) & 2.22 (0.05) & 2.16 (0.06) \\
\bottomrule
\end{tabular}
}
\end{table*}

\section {Results}

This section presents the performance of our proposed MVI method compared to 11 baseline methods under varying missing value types and rates. 

\subsection{Effects of missing value types}

We present the performance rank order of the imputation methods in Table~\ref{tab:rank_order_table} because no model can perform the best on all tabular data sets. The results specific to three missing value types are discussed in the following.

\subsubsection{MCAR missing type}

Table~\ref{tab:imputation_mcar} presents the mean NRMSE scores of the imputation methods on MCAR-type data. As expected, the median value imputation performs the worst as the single imputation method. KNN shows promising results for missing rates 30\% and 50\%, outperforming MICE in terms of overall rank. MICE outperforms missForest in eight data sets when the missing rate is 10\%. In contrast, missForest outperforms MICE and KNN for ten and eight data sets with 30\% missing values, respectively. When the missing rate is 50\%, missForest outperforms MICE and KNN on eight and seven data sets, respectively. Therefore, missForest is generally superior to MICE- and KNN-based imputation methods. Among deep learning methods, Diffputer outperforms GAIN in seven data sets at a 10\% missing rate. However, GAIN performs better than Diffputer in 10 data sets when the missing rate is 50\%. In contrast, attention-based methods outperform GAIN and Diffputer in all missing data types and missing rates. The proposed DeepIFSAC achieves the best NRMSE scores for six data sets with missing rates of up to 50\%. A closer look reveals that DeepIFSAC outperforms imputation with between-feature attention in seven and six data sets with 10\% and 30\% missing rates, respectively. Notably, the overall performance of attention-based imputation methods, including DeepIFSAC, is negatively impacted by data sets with IDs 187 and 40982. The average rank order presented in Table~\ref{tab:rank_order_table} shows that DeepIFSAC is the best imputation method for missing rates up to 50\% but overall the best for MCAR-type data, as shown in Figure~\ref{fig:mcar_avg_rank}. 
 
\subsubsection{MAR missing type}
Table~\ref{tab:imputation_mar} presents the performance of the imputation methods for MAR-type data. As usual, single-value imputation using median values performs the worst. MICE outperforms missForest and KNN in seven and nine data sets with 10\% missing values, respectively. In contrast, missForest outperforms MICE and KNN for eight and seven data sets with 50\% missing values, respectively. This suggests that the non-linear missForest method is superior to its linear counterpart (MICE) at higher missing rates. Among deep learning-based imputation methods, GAIN outperforms Diffputer for missing rates ranging from 30\% to 90\% in Table~\ref{tab:rank_order_table}. However, attention-based methods outperform GAIN and Diffputer regardless of the missing rate. Our proposed DeepIFSAC is the best imputation method for MAR-type data with missing values up to 50\%. DSAN and attention between samples show competitive performance when the missing rate is 70\% or higher. However, DeepIFSAC ranks the best when averaged across all missing rates, as shown in Figure~\ref{fig:mar_avg_rank}. In contrast, attention-based methods perform poorly on data sets with IDs 187, 1050, and 40982. missForest outperforms attention learning in data sets with IDs 1049, 1050, and 40982. 


\subsubsection{MNAR missing type}

Table~\ref{tab:imputation_mnar} presents the imputation performance for MNAR-type data. Median value imputation performs worse than KNN, MICE, and missForest for all missing rates. KNN shows promising results by outperforming MICE and missForest in six and ten data sets with 10\% missing values, respectively. However, MICE and missForest perform equally well on six data sets each when the missing value rate is 50\%. The Diffputer outperforms GAIN in seven and eight datasets when the missing rates are 30\% and 50\%, respectively. The average order of rank presented in Table~\ref{tab:rank_order_table} and Figure~\ref{fig:mnar_avg_rank} shows that DeepIFSAC is the best among all methods from 10\% to 90\% missing rates.

\subsection{Effects of data types and missing rates}

Twelve OpenML tabular data sets are evenly grouped into easy and hard categories based on the contrast between linear and non-linear classifier performance. Eighteen experimental scenarios, comprising six data sets and three missing value types, are investigated for each data category. In the easy category,  the proposed DeepIFSAC method performs the best in seven out of 18 scenarios with 10\% missing rates. At a 50\% missing rate, DeepIFSAC outperforms all other imputation methods in eight experimental scenarios, while missForest ranks the best in six. In the hard data category, DeepIFSAC performs best in 11 of 18 scenarios with a 10\% missing rate and 9 of 18 scenarios with a 50\% missing rate. Therefore, DeepIFSAC wins the highest number of experimental scenarios compared to any other model for these missing rates. 


\begin{table*}[t]
\centering
\caption{Mean F1 scores in classifying imputed All of Us~\citep{allofus} heart failure patient data.}
\label{tab:gbt_f1_allofus}
\scalebox{0.5}{
\begin{tabular}{lcccccccccc}
\toprule
 \shortstack{Median\\[0.2em]} &  \shortstack{KNN\\[0.2em]} & \shortstack{MICE\\~\citep{mice}} & \shortstack{missForest\\~\citep{missForest}} & \shortstack{DSAN\\~\citep{lee2023}} & \shortstack{MIDASpy\\~\citep{Midaspy_Lall_2022}} & \shortstack{Between-\\feature~\citep{kowsar2024}} & 
 \shortstack{Between-\\sample~\citep{kowsar2024}}& 
 {\shortstack{DeepIFSAC\\(Proposed)}} \\
 \midrule
 76.15 (9.26) & 83.59 (1.04) & 85.10 (1.48) & 80.30 (3.03) & 82.29 (2.12) & 86.04 (0.6) & 86.59 (1.44) & 87.24 (1.24) & 
 {\bf 87.49 (1.35)} \\
\bottomrule
\end{tabular}
}
\vspace{-10pt}
\end{table*}

\subsection{Imputation of health records data}

Missing values are inevitable in medical records, where ground truth values are unknown. NRMSE scores cannot be obtained without ground truth values to evaluate the imputation performance. An indirect approach is to perform classification using the imputed data and infer imputation performance from classification performance.   A data set is obtained from the All of Us EHR by selecting two cohorts of patients based on the diagnostic codes of the International Classification of Diseases (ICD-10): one with a diagnosis of heart failure and the other without events of cardiovascular disease. Table~\ref{tab:gbt_f1_allofus} shows the F1 scores of a gradient-boosting tree (GBT) classifier following the five-fold cross-validation of various model-imputed data. The MICE method appears to be superior to median value imputation, KNN, missForest, and one of the deep learning imputation methods (DSAN). Notably, MICE is the most popular method in biostatistics used to impute health records data~\citep{Samad2019Jaac}. All attention learning methods, including the proposed DeepIFSAC, produce better classification performance than baseline imputation methods. In particular, a 1\% improvement in classification performance implies the correct diagnosis of an additional 20 patients with heart failure in a sample size of 2000, which is of significant value in clinical practice. 

\subsection{Computation time and ablation studies}
\begin{table*}[t]
\centering
\caption{Comparison of computation time in seconds for training the imputation methods for 1000 epochs in five-fold cross-validation.}
\label{tab:time_complexity}
\scalebox{0.57}{
\begin{tabular}{lccccccccccc}
\toprule
 \shortstack{Data\\set} &  \shortstack{Median\\[0.2em]} &  \shortstack{KNN\\[0.2em]} & \shortstack{MICE\\~\citep{mice}} & \shortstack{missForest\\~\citep{missForest}} & 
\shortstack{GAIN\\~\citep{yoon2018gain}} & \shortstack{Diffputer\\~\citep{diffimpute}} & \shortstack{DSAN\\~\citep{lee2023}} & \shortstack{MIDASpy\\~\citep{Midaspy_Lall_2022}} & \shortstack{Between-\\feature~\citep{kowsar2024}} & 
 \shortstack{Between-\\sample~\citep{kowsar2024}}&{\shortstack{DeepIFSAC\\(Proposed)}} \\
\midrule
37   & 4  & 23  & 15   & 214 & 45 & 89 & 232   & 330   & 410   & 440 & 493 \\
1497 & 33 & 203 & 190  & 1195  & 514 & 489 & 7172 & 6231  & 4758  & 5293  & 5309 \\
\bottomrule
\end{tabular}
}
\end{table*}

We compare the computational times of the imputation methods in Table~\ref{tab:time_complexity} using two representative data sets. Median value imputation, KNN, MICE, and missForest are non-deep learning approaches without training epochs. All deep imputation methods are trained on four data folds for 1000 epochs and tested on the left-out fold to obtain the time required for training and inference. As expected, deep learning models require more computational time than traditional methods to yield substantially better imputation performance than non-deep imputation methods. However, the computational time requirements of deep imputation methods vary, depending on the number of training epochs and the size of the data set. Although the proposed DeepIFSAC method needs less computational time for a larger and hard-to-classify data set (ID 1497) compared to the deep imputation baselines, the opposite trend is observed for a smaller and easier-to-classify data set (ID 37).

\begin{table*}[t]
\centering
\caption{Ablation study on augmentation strategies and contrastive learning in DeepIFSAC under MCAR across three missing rates.}
\label{tab:ablation}
\scalebox{0.6}{
\begin{tabular}{cccccc}
\toprule
{\shortstack{Dataset\\[0.2em]}} & {\shortstack{Missing Rate (\%)\\[0.2em]}} & {\shortstack{DeepIFSAC \\without CutMix}} & {\shortstack{DeepIFSAC \\Zeros replace CutMix}} & {\shortstack{DeepIFSAC \\ without Contrastive}} & {\shortstack{DeepIFSAC \\(Proposed)}} \\
\midrule
     & 10 & 0.35 (0.03) & 0.33 (0.04) & 0.18 (0.05) & 0.17 (0.05) \\
37     & 30 & 0.58 (0.02) & 0.58 (0.01) & 0.30 (0.06) & 0.29 (0.06) \\
     & 50 & 0.72 (0.04) & 0.74 (0.03) & 0.35 (0.05) & 0.34 (0.05) \\
\midrule
   & 10 & 0.31 (0.01) & 0.21 (0.01) & 0.18 (0.01) & 0.18 (0.00) \\
1497   & 30 & 0.52 (0.01) & 0.38 (0.01) & 0.32 (0.01) & 0.32 (0.01) \\
   & 50 & 0.69 (0.01) & 0.51 (0.01) & 0.44 (0.01) & 0.43 (0.01) \\
\bottomrule
\end{tabular}
}
\vspace{-10pt}
\end{table*}

Table~\ref{tab:ablation} presents an ablation study to show the effect of data augmentation and contrastive learning (Equation\ref{eq:infoNCE_loss}) during model training. CutMix data augmentation method improves the imputation performance compared to the case without augmentation or using zero values for data augmentation similar to ~\citep{lee2023, Midaspy_Lall_2022, aimnet}. The proposed DeepIFSAC method produces a better imputation performance than its counterpart without contrastive learning. These results are consistent for both representative data sets, suggesting the value of joint attention learning within the contrastive framework. Furthermore, we examine our model's sensitivity to different hyperparameter settings. We vary the embedding size (16, 32, 64, 128), the number of attention heads (4, 6, 8, 10), and the learning rates (1e-4, 3e-5, 3e-4). Across all these settings, the NRMSE scores of the proposed DeepIFSAC change at the third decimal point, indicating the model's robustness to hyperparameter changes.

\section{Discussion of the results}

This paper uses between-feature and between-sample attentions within a contrastive learning framework to perform missing value imputation. The experimental results in the previous section reveal several key findings. First, the proposed DeepIFSAC method yields the best overall imputation performance, especially with MAR- and MNAR-type values. Notably, MAR and MNAR are more challenging and realistic missing values than the MCAR type. Second, the CutMix method improves the MVI performance compared to the current method of random masking and filling with zeros~\citep{MIDA_Gondara2018,lee2023, Midaspy_Lall_2022,aimnet}. Third, attention between samples offers complementary value in imputation when the missing rate is high or missing values do not depend on features. The proposed DeepIFSAC method leverages the complementary values of between-feature and between-sample attention to outperform individual attention-based imputation. Fourth, the contrastive learning framework improves the joint attention between features and between samples in missing value estimation. Fifth, attention-based imputation is superior to other deep learning approaches, including the denoising autoencoder, GAN, and diffusion-based methods.  The following sections provide insight and further explanations of our research findings.

\subsection {Comparison with the literature}
Sun et al. have recently conducted an extensive survey of missing value imputation methods~\citep{sun2023deep}. They use ten data sets to conclude that deep-generative models (e.g., GAIN and Variational Auto-Encoder (VAE)) underperform traditional models (MICE, MissForest) unless the sample size is large ($n > ~30,000$). In line with our study, they report the superiority of MICE and missForest over GAN- and VAE-based imputations for missing rates from 10\% to 50\% and three missing value types. Although the survey demonstrates the limitation of deep imputation methods, it does not study any attention-based imputation. Chen et al. ~\citep{rethinking_diff} report the superiority of GAN-based imputation over its diffusion counterpart after comparing GAIN with MissDiff and CSDI\_T~\citep{tabcsdi}. In contrast, we argue and demonstrate that the row-column structure of tabular data is most effectively and intuitively imputed by considering row-column attention.

The discrepancy in model performance in the literature can be attributed to the variability in batch size, data set selection, and the number of training epochs. For example, the Diffputer article uses a batch size of 4096, which is much larger than the standard batch size like ours (128). GAIN and Diffputer authors train their models for 10,000 epochs, unlike our experiments, training for only 1,000 epochs. Unlike other studies, we benchmark all methods without the assumption or requirement of large sample or batch sizes or longer training epochs for convergence.

\subsection{Effects of missing value types}\label{subsec:miss_value_type}

The effect of missing value types can be related to the design of the imputation model for an explanation. In MAR-type data, missing values depend on values of other features. This may explain why feature regression-based models, such as MICE and missForest, effectively impute MAR-type data. Attention between features may effectively impute MAR-type data for the same reason. In contrast, MNAR-type data have a missing value due to itself, independent of other features. The absence of feature dependence negatively impacts the performance of regression-based models (missForest) or between-feature attention. In this scenario, the attention between samples performs better because it learns the missing values from other samples instead of features. Generative models often rely on cross-feature relationships and diffusion-based imputations ~\citep{ouyang2023missdiff,Diffputer} usually exclude MNAR-type data. This suggests the need to develop imputation methods tailored to data and missing-value types.

\subsection{Effects of data characteristics}\label{subsec:data_specific}
The heterogeneity of tabular data sets, unlike image or text data, has some data-specific effects on model performance because there is no single best model for all data sets. An explanation is necessary to understand why several data sets (IDs 187, 1049, 1050, and 40982) are better imputed by traditional methods than the proposed attention-based imputations. The data set with ID 187 has the lowest sample size (178). Deep learning methods require large sample sizes for optimal performance, which may explain the underperformance of attention-based methods on this data set.  Data sets with IDs 1049 and 1050 are under the hard-to-classify category and have similar sample sizes and feature dimensionality. Unlike other data sets, the features of these two data sets are of large value with very high standard deviations. The best performance of missForest on the MCAR and MAR types of these data sets may be because random forest models are insensitive to feature scales, unlike deep learning methods~\citep{Grinsztajn2022}. Data sets with IDs 1049, 1050, and 40982 have the highest feature dimensions within the hard-to-classify category. High feature dimensions in certain tabular data sets may affect the performance of attention-based imputation methods.

\subsection{Resilience to high missing rates}

Our results reveal some effects of higher missing rates on imputation performance. Such effects have largely been unexplored in the literature. Most deep imputation methods limit benchmarking to low missing rates, up to 30\% \citep{lee2023, aimnet}. Some features with predictive values can be missing at a high rate because their importance has not yet been discovered in the domain. Excluding such features from predictive analytics limits the performance of data-driven models.
For example, the second most important predictor of all-cause mortality in \citep{Samad2019Jaac} has more than 50\% missing values. Therefore, an imputation model robust to high missing rates is a valuable contribution. The worst approach, median value imputation, is the least sensitive to high missing value rates. This is because median values do not depend on or learn from observed values as strongly as other sophisticated models. Attention between samples is robust at high missing rates (70\% and 90\%) when the properties between features are compromised in MAR- and MCAR-type data. 


\subsection {Attention versus other deep imputations}

 The superior performance of attention-based imputations, including our proposed DeepIFSAC method, may be explained as follows. A missing value lies at the intersection of a row and a column of a tabular data set. Therefore, attending to columns and rows learns complementary estimates for missing values, unlike individual attention mechanisms. Without an association between features, attention between samples aids in missing value estimation from similar samples. The ablation study shows that the CutMix data augmentation improves the imputation performance of our proposed DeepIFSA method. One reason for this improvement is that CutMix uses actual feature values from another sample to initialize the masked values instead of filling them with zeros in other imputation methods~\citep{Midaspy_Lall_2022, lee2023}. CutMix has shown the best representation of tabular data in classification tasks using a contrastive learning framework~\citep{scarf, rabbani2024}.

In contrast, deep-imputation solutions based on GAN, autoencoders, and diffusion-based approaches may underperform attention-based imputations for several reasons. The autoencoder reconstructs missing values from the projection of features in encoding units. NRMSE scores tend to explode when the denoising autoencoder (MIDASpy method) imputes certain data sets~\citep{aimnet}, similar to our findings. The performance of an autoencoder can be affected due to improper weight initialization and the vanishing gradient problem, particularly in deeper architectures~\citep{Li2018Autoencoder, Li2021EncoderDecoder}. These issues can lead to unstable training and increased reconstruction errors. Diffusion-based methods make a generalized assumption about all feature distributions (e.g., Gaussian distribution). An assumption about the feature distribution may not be accurate, and distribution matching may not guarantee alignment between actual and estimated values. GAN-based models, such as GAIN, do not impose such an assumption and often outperform diffusion-based imputations~\citep{ouyang2023missdiff, diffimpute}. However, GAN-based models generate values from random noise, which may align with missing completely random values (e.g. MCAR). In general, diffusion- and GAN-based models are not designed to learn row-column relationships in line with missing value types that are not completely random (e.g., MAR or MNAR). In contrast, our proposed attention-based approach yields superior performance by learning row-column attentions to support different missing value types and rates.

\subsection {Limitations and future work}

Despite its superior performance against state-of-the-art methods, the proposed DeepIFSAC method has some limitations. The high computational time of the model may dissuade some practitioners from adopting deep imputation methods. However, data imputation happens once in a lifetime to ensure high-quality data in downstream data analytics and predictive modeling. The extra time spent on deep imputation will provide better quality data than traditional imputation methods to support more reliable data-driven decision-making. The proposed method is designed to impute numerical features by data reconstruction and is evaluated using NRMSE scores. In contrast, the imputation of categorical features requires a separate model with a classifier head evaluated by a classification score (e.g., F1 score) instead of NRMSE scores. The model will need an extension to support mixed-type data imputation. The proposed DeepIFSAC performs worse with limited sample sizes ($<200$) than traditional imputation methods because deep learning models demand large sample sizes. Attention learning can suffer from out-of-memory problems with high-dimensional data ($>80$), which is alleviated by joint contrastive learning~\citep{rabbani2024}. The dimension of most tabular data in practice is relatively smaller than that limit.  Despite these limitations, the proposed DeepIFSAC method performs best for various tabular data sizes and domains, including data from real-world health records.

\section {Conclusions}

This paper combines between-feature and between-sample attention to attend to rows and columns of tabular data to impute missing values in tabular data. The CutMix data augmentation method, within its native contrastive learning, improves the imputation accuracy of the combined between-feature-sample attention learning. The proposed imputation algorithm is effective across varying missing value types and rates and against state-of-the-art methods. The superiority of the proposed imputation method is also reflected in the classification performance of the imputed data from real electronic health records. Despite the superior performance rank, a data set-specific consideration is still needed to identify the most effective imputation method considering the heterogeneity of tabular data sets.

\bibliographystyle{elsarticle-num}
\bibliography{myBib}

\newpage 
\appendix
\section{Additional Experimental Results}

\begin{figure}[H]
\centering
\includegraphics[trim=0cm 0cm 0cm 0cm, width=0.9\textwidth] {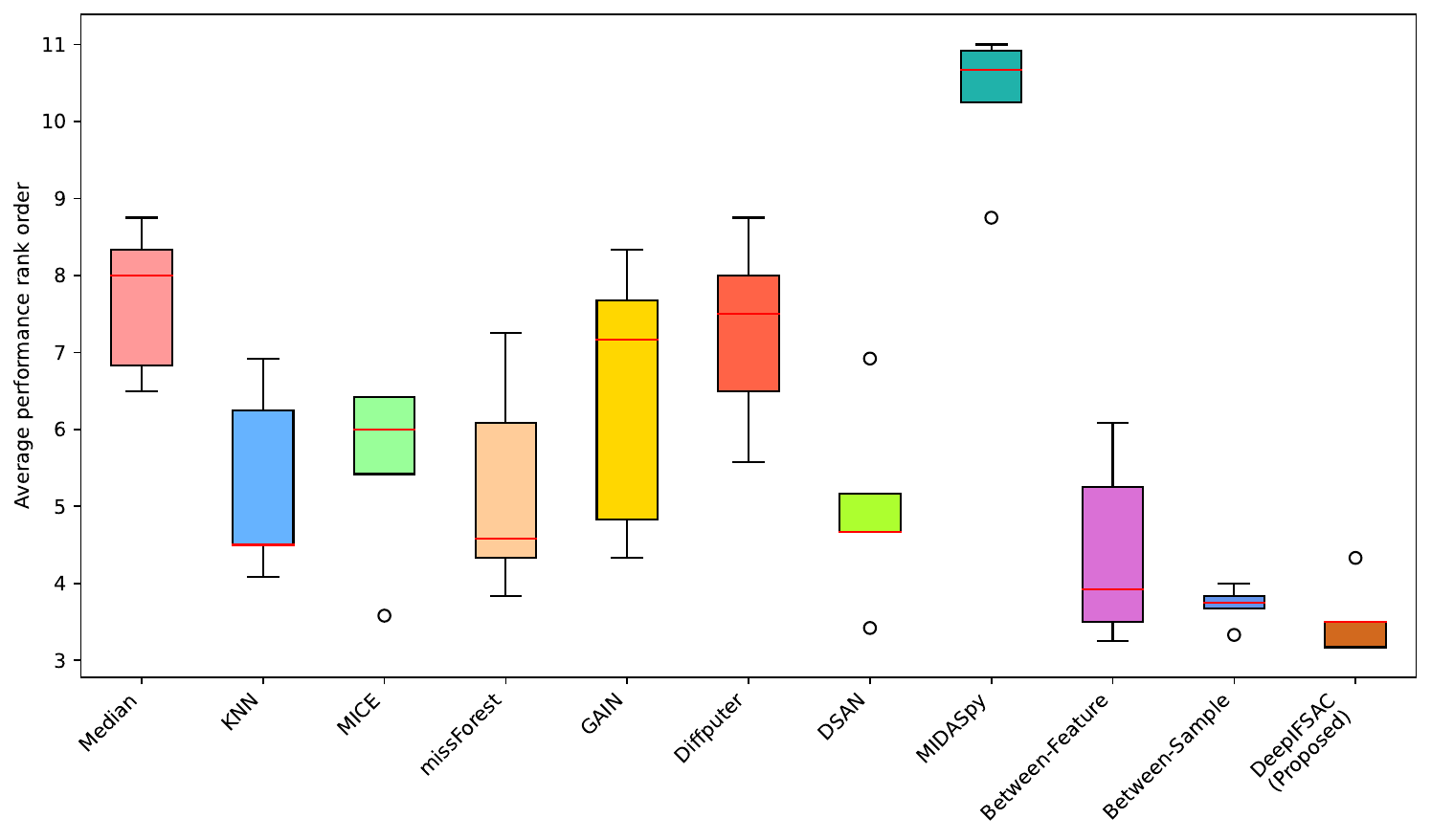}
\vspace{-10pt}
\caption {Average rank order of individual imputation methods obtained across 10\% to 90\% MCAR type values.}
\label{fig:mcar_avg_rank}
\end{figure}

\begin{figure}[b]
\centering
\includegraphics[trim=0cm 0cm 0cm 0cm, width=0.9\textwidth] {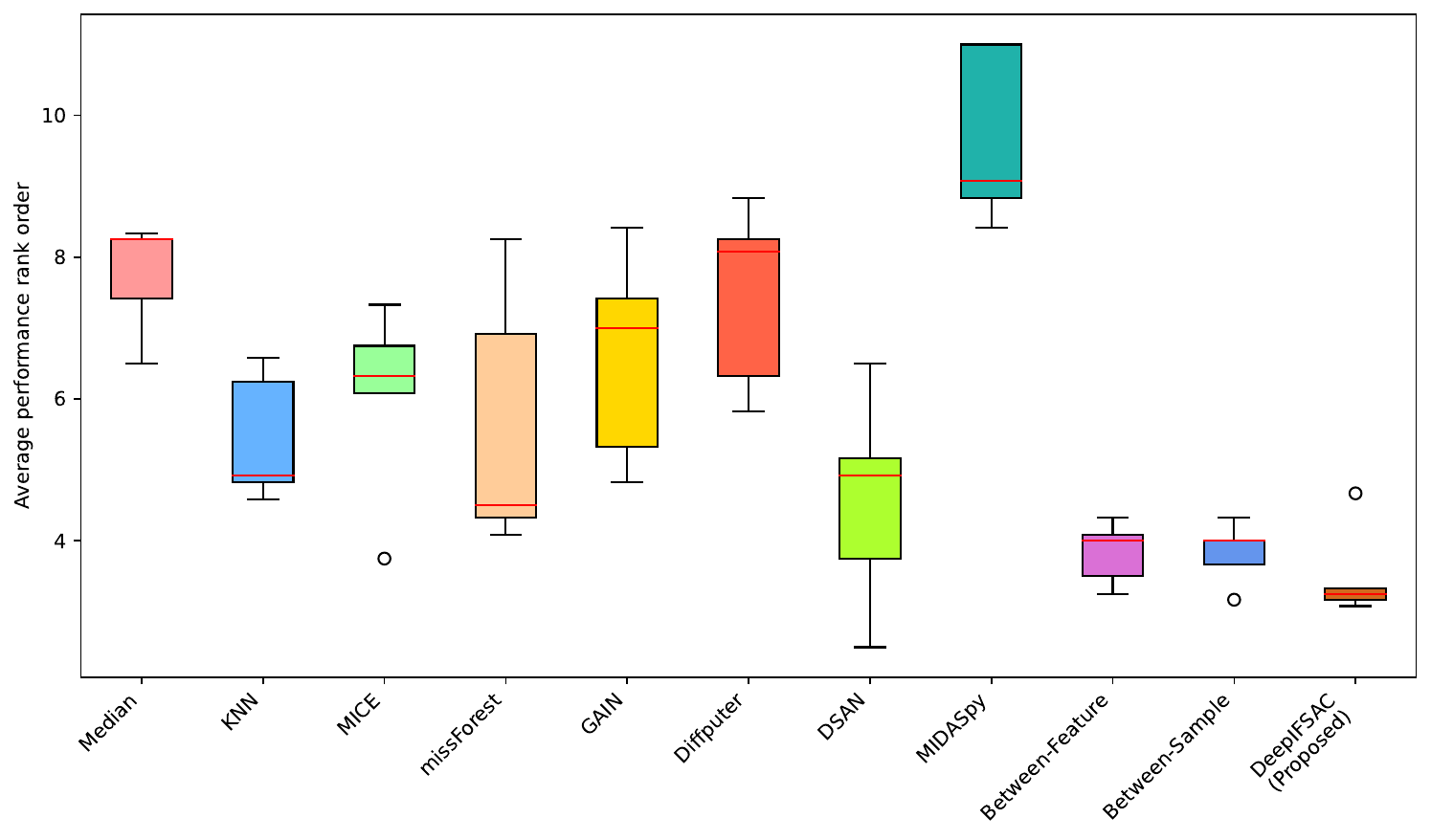}
\vspace{-10pt}
\caption {Average rank order of individual imputation methods obtained across 10\% to 90\% MAR type values.}
\label{fig:mar_avg_rank}
\end{figure}

\begin{figure}[t]
\centering
\includegraphics[trim=0cm 0cm 0cm 0cm, width=0.9\textwidth] {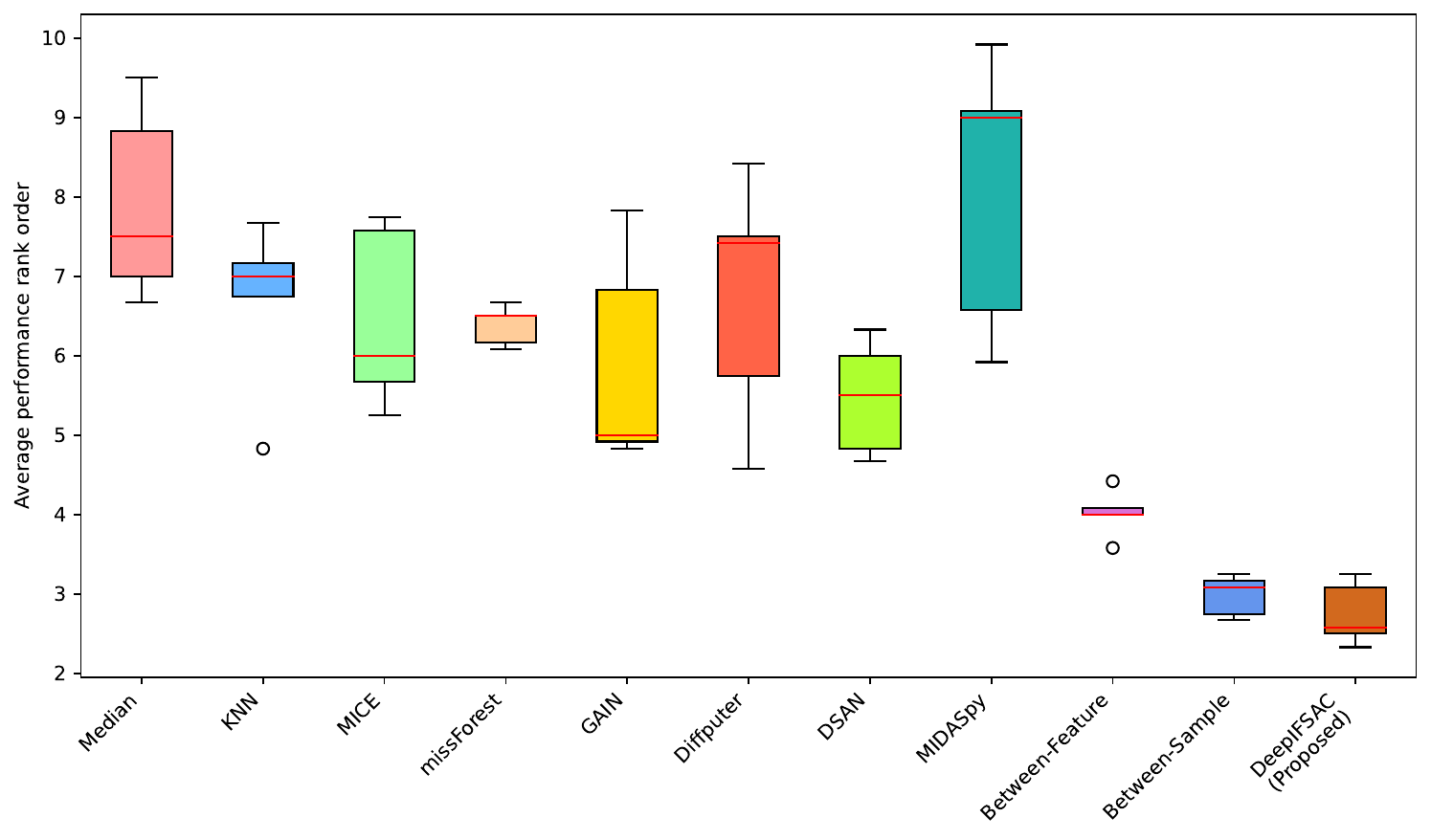}
\vspace{-10pt}
\caption {Average rank order of individual imputation methods obtained across 10\% to 90\% MNAR type values.}
\label{fig:mnar_avg_rank}
\end{figure}

This section provides additional results demonstrating the average rank order of all imputation methods for the three missing data types. Missing Completely at Random (MCAR) in Figure \ref{fig:mcar_avg_rank}, Missing at Random (MAR) in Figure \ref{fig:mar_avg_rank}, Missing Not at Random (MNAR) in \ref{fig:mnar_avg_rank} show the effect of different missing value types. The average rank in each figure is obtained across the 10\% to 90\% missing values to demonstrate resilience to varying missing rates.

\label{app:exp_results}

\end{document}